\documentclass[journal]{IEEEtran}
\usepackage{cite}
\usepackage{array}
\usepackage{mdwmath}
\usepackage{mdwtab}
\usepackage{eqparbox}
\usepackage{subfigure,graphicx}
\usepackage{enumerate}
\usepackage{epstopdf}
\usepackage{multirow}
\usepackage{algorithm, algpseudocode, algorithmicx}
\usepackage{flushend}
\usepackage{color}
\usepackage{epstopdf}

\newcommand{\tabincell}[2]{\begin{tabular}{@{}#1@{}}#2\end{tabular}}

\begin{document}

\title{{Robust Gait Recognition by Integrating Inertial and RGBD Sensors}}

\author{Qin~Zou,
        Lihao~Ni,
        Qian~Wang,
        Qingquan~Li,
        and Song Wang
%
\thanks{Q.~Zou, L.~Ni, Q.~Wang 
are with
School of Computer Science, Wuhan University, Wuhan 430072,
P.R.~China (E-mails: \{qzou, lhni, qianwang
\}@whu.edu.cn).}
\thanks {Q.~Li is
with Shenzhen Key Laboratory of Spatial Smart Sensing and Service,
Shenzhen University, Guangdong 518060, P.R.~China (E-mail:
liqq@szu.edu.cn). }
\thanks {S.~Wang is with
Department of Computer Science and Engineering, University of South
Carolina, Columbia, SC 29200, USA (E-mail: songwang@cec.sc.edu).}

}




\maketitle


\begin{abstract}
Gait has been considered as a promising and unique biometric
for person identification. Traditionally, gait data are collected
using either color sensors, such as a CCD camera,  depth sensors,
such as a Microsoft Kinect, or inertial sensors, such as an accelerometer.
However, a single type of sensors may only capture part of the dynamic
gait features and make the gait recognition sensitive
to complex covariate conditions, leading to fragile gait-based person
identification systems. In this paper, we propose to combine all three
types of sensors for gait data collection and gait recognition, which
can be used for important identification applications, such as
identity recognition to access a restricted building or area.
We propose two new algorithms, namely EigenGait and TrajGait, to extract gait features
from the inertial data and the RGBD (color and depth) data, respectively. Specifically,
EigenGait extracts general gait dynamics from the accelerometer readings
in the eigenspace and TrajGait extracts more detailed sub-dynamics by analyzing
3D dense trajectories. Finally, both extracted features are fed into a
supervised classifier for gait recognition and person identification.
Experiments on 50 subjects, with comparisons to several other state-of-the-art
gait-recognition approaches, show that the proposed approach
can achieve higher recognition accuracy and robustness.
\end{abstract}


\IEEEpeerreviewmaketitle

\section{Introduction}

\IEEEPARstart{U}{sing} gait, or the manner of walking, for person identification
has been drawing more and more attention in recent years~\cite{Nixon2006,zhang2010tsmc,ding2015tc},
due to its capability to recognize a person at a longer distance than
the traditional biometrics based on face, fingerprint and iris recognition.
However, in practice gait biometrics usually suffer from two issues. First,
the data collected by a single type of sensors, \textit{e.g.}, a CCD camera, may only capture
part of the gait features and this may limit the gait recognition accuracy.
Second, gait biometrics are usually sensitive to hard-covariate conditions, \textit{e.g.},
walking with hands in pocket or with loadings. In this paper, we
propose to combine gait data collected by different types
of sensors to promote the gait recognition accuracy and the robustness.

In the previous research, three types of sensors have been
used for gait data collection and gait recognition --
color sensors, depth sensors and inertial sensors.
Using color sensors, \textit{e.g.}, CCD cameras, a
walking person can be captured into a video, in which
each frame is a 2D RGB (color) image of the person and the surrounding
environment. Gait recognition on such a  video
is usually achieved by segmenting, tracking, and analyzing the silhouette of the walking person
on each frame~\cite{Kale02fg,Han03cvpr,Liu06pami,ran2010tsmc,gu2010tsmc,Kusa13tifs,Boulgouris13tip,goffredo2010tsmc}.
The silhouette segmentation and tracking can be difficult when
the color of the person is similar to the color of the surrounding environment in the video.
In addition, color sensors generally capture the dynamic gait features in a 2D space.

Using depth sensors, such as the line-structure light devices, it is usually
easier to segment a walking person from the
surrounding environment, when there is no other moving objects around.
In addition, from the depth data, 3D dynamic
gait features can be derived for gait recognition~\cite{Sivapalan11ijcb,John13icip,Chatto14jvcir}.
However, in practice depth data may contain noise and
errors, especially at the spots with strong reflectiveness, \textit{e.g.},
on a reflective clothing, where the depth value is totally invalid.
Such errors may lead to incorrect gait features and gait recognition
results.

Different from color and depth sensors, which are installed
to capture the walking person at a distance to collect gait data,
inertial sensors such as accelerometers and gyroscopes collect
gait data by attaching to and moving with the
person~\cite{Mantyjarvi05iccasp,Liu07icbb,Kwapisz10icb,XuCMU12btas,Trung11ijcb,Trung12icb,Sun14sensors}.
The inertial-sensor based gait recognition mainly
benefits from the extensive use of smart phones --  people
always carry their smart phones and almost all the smart phones
have integrated inertial sensors of accelerometers and gyroscopes.
Considering the usability, the smart phone must be allowed to be
placed in any pockets with different orientations
when we use its inertial sensors for gait recognition.
Such different placements and orientations of the sensors may
vary the inertial data and affect the gait recognition accuracy~\cite{XuCMU12btas}.

In general, each type of the above-mentioned sensors can capture part of
the gait features with different kinds of errors and incompleteness.
For example, depth and inertial sensors capture 3D gait features and color
sensors capture 2D gait features. Meanwhile, the inertial data, such as the accelerometer
readings, portrait the motion pattern of the whole body and provide a
general description to the gait dynamics, while the color and depth
data can be used to infer the motion of many body parts and
provide more detailed sub-dynamics of the gait. It is natural
to assume that the gait features derived from different sensors can
complement each other. This motivates the proposed approach
to integrate the color, depth and inertial sensors for
more accurate gait recognition.

Sensitivity to complex covariate conditions is another main issue
in gait biometrics~\cite{Liu06pami}. For example, gait data from a sensor may
look  different when the same person walks with hands in pocket or
with loadings. Such a difference increases the variance
of a person's gait features and reduces the gait recognition
accuracy. In this paper, through carefully designed
experiments, we show that the proposed approach of integrating
different sensors can also improve the
robustness of gait recognition under complex covariate conditions.

\begin{figure}[!t]
    \centering
    \centerline{\includegraphics[width=0.99\linewidth]{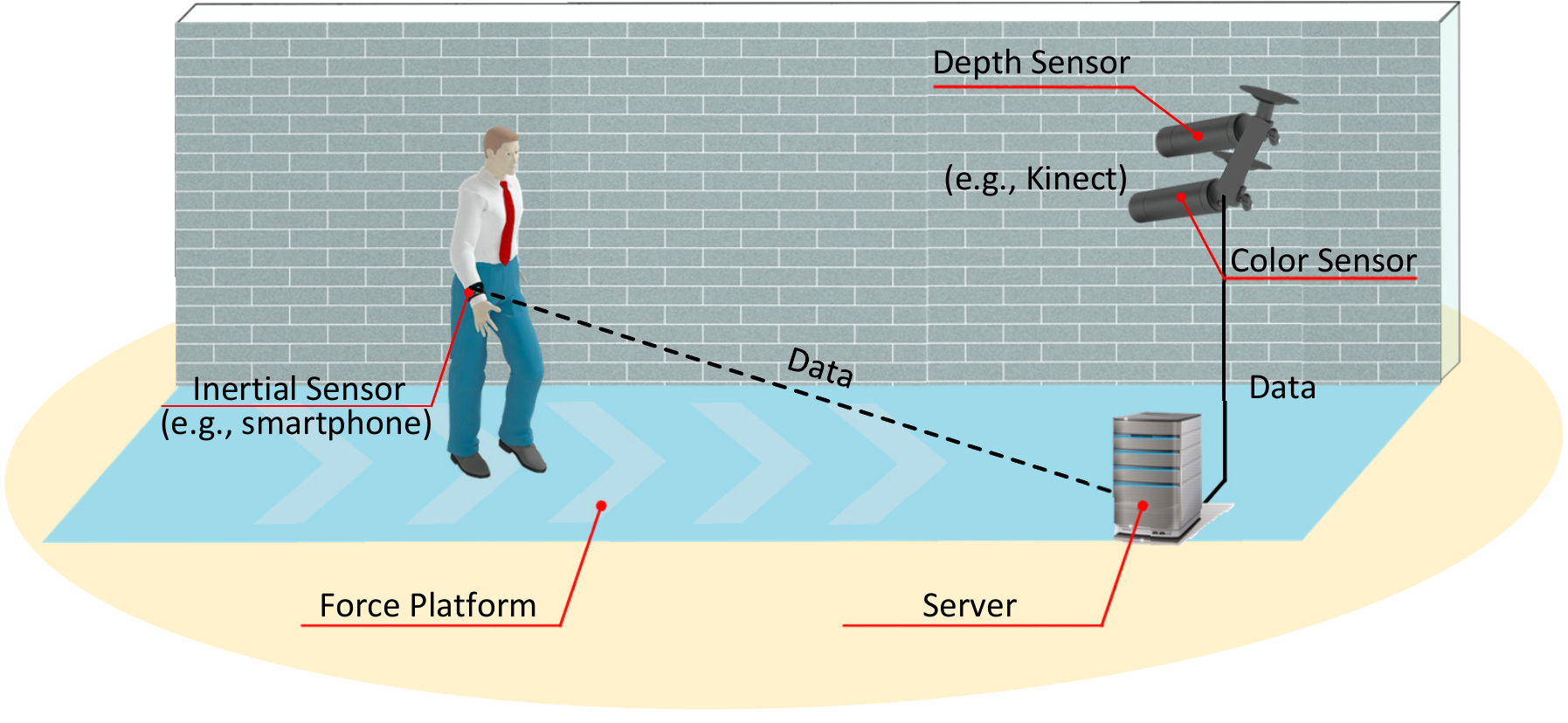}}
    \caption{{The application scenario of the proposed approach: a gait-based person identification system for accessing a restricted area. }}
    \label{fig:usage}
\end{figure}

As a practical application scenario, the proposed approach of
integrating different sensors for gait recognition can be used
for person identification to access a restricted area or building.
As illustrated in Fig.~\ref{fig:usage}, at the entrance of a restricted
area, a user simply walks on a force platform
to get his identity verified. During his walk, a pre-installed
client application in his smart phone sends real-time
inertial-sensor readings to the server by wireless communication.
At the same time, color and depth sensors, mounted over the ceiling
and facing the platform, collect the RGBD (color and depth) data
and send them to the server. In the server, the proposed approach
can integrate all the data and perform gait recognition to identify
whether he is an authorized user or not. Other than a higher gait
recognition accuracy,  such an identification system also has
good security -- even if the smart phone is hacked to send
forged inertial data to the server, it is difficult to forge the
RGBD data since color and depth sensors are not controlled by
the user.

Following the scheme of identification illustrated in Fig.~\ref{fig:usage}, in this paper
we use accelerometer in the smart phone to collect inertial data
and Microsoft Kinect to collect the RGBD (color and depth) data.
We develop a new EigenGait algorithm to capture the general
gait dynamics by analyzing the inertial
data in the eigenspace and a new TrajGait algorithm to
capture more detailed gait sub-dynamics based on
the 3D trajectories extracted from the RGBD video.
The extracted features on general dynamics and sub-dynamics of gait are then
integrated and fed into a supervised classifier for gait recognition
and person identification. In the experiments, we collect three sets of
inertial and RGBD data from 50 subjects and evaluate
the proposed approach under various covariate conditions.
Comparison results with other approaches confirm that
the gait recognition accuracy and robustness can be improved by
integrating different types of sensors.
The main contributions of this paper lie in four-fold.
\begin{enumerate}[    $\vcenter{\hbox{\tiny$\bullet$}}$]
\item First, a multi-sensor integration method is proposed for gait recognition, in which inertial sensor, color sensor and depth sensor are integrated to capture gait dynamics. The multi-sensor data fusion leads to more robust gait-recognition performance.

\item Second, an EigenGait algorithm is developed to describe the general
gait dynamics by analyzing the time-series acceleration data in the eigenspace. The
extracted features are more effective than that produced by Fast Fourier Transforms (FFT) or Wavelet
Transforms.

\item Third, a TrajGait algorithm is proposed to describe the detailed
sub-dynamics of gait by analyzing the RGBD videos.
In TrajGait, 3D dense trajectories are derived from the RGBD videos and used
for representing the gait features. We found that such gait features are
more discriminative than the depth- or skeleton- based features in gait recognition.

\item Finally, three new datasets, with both RGBD and accelerometer data,  are collected on 50 subjects. They
can be used to quantitatively evaluate and compare the performance of different gait recognition methods.
\end{enumerate}


The remainder of this paper is organized as follows.
Section~\ref{sec:relate} reviews the related work.
Section~\ref{sec:method}  introduces the proposed approach, including sensor setting,
data collection, gait feature extraction, and integrated gait recognition.
Section \ref{sec:experiment} reports the experiments and results.
Section~\ref{sec:conclusion} concludes our work and briefly discuss the possible future work.


\section{Related Work}\label{sec:relate}
The ideas and experiments of gait recognition can be traced back
to Cutting and Kozlowski's work~\cite{Cutting77}, in which the
manner of walking, \textit{i.e.}, the gait, was found to be possible
to identify a person. Since then, gait-based person identification
has attracted extensive attention in both academia and
industry~\cite{Herran14sensors}, and a number of gait recognition
methods have been proposed. In these methods, three types of sensors
are mainly used for gait data collection, namely the color sensor,
the depth sensor, and the inertial sensor. Hence, the gait
recognition methods can be classified into the color-based, the
depth-based, and the inertia-based. In this section, we briefly
overview them, as well as a brief overview to other action-based
biometrics.

\noindent
\textbf{Color-based methods}. The color-based methods had
a rapid development in the early
days~\cite{Kale02fg,BenAbdelkader02fg,wang03pami,Han03cvpr,Kale04tip,Liu06pami,Tao07pami,wang03iccv,Tolliver03avb,Han04cvpr,Zhang05neurcomp}.
These methods can be classified into the model-free methods and the
model-based methods. In the model-free methods, gait features are
often extracted by analyzing the shapes, or contours, of the
silhouettes in successive frames. In addition, features on the
velocity, texture and color are also examined. One important work
among them is the GEI (gait energy image) method~\cite{Han03cvpr},
which represents gait dynamics by an aligned and normalized
silhouettes over a gait cycle. The GEI provides a compact
representation of the spatial occupancy of a person over a gait
cycle. However, partitioning gait cycles from a color video is
rarely easy. In~\cite{wang03pami,BenAbdelkader01AVPA}, silhouettes
were produced by background subtraction, and gait features were
extracted by principal component analysis. In~\cite{chen2011pr}
and~\cite{venkat2011ijcv}, statistic methods were employed to
analyze the gait characteristics on a sequence of binary silhouettes
images. Motion has been exploited for gait
representation~\cite{Hu13tc,lam2011pr,Castro14icpr}.
In~\cite{Hu13tc}, motion is described by local binary patterns,
and HMM (Hidden Markov Model) is then applied to distinguish the gait dynamics of different persons.
In~\cite{Kusa14ivc}, gait motions were encoded based on a set of
spatio-temporal interest points from a raw gait video. These
interest points were detected by using Harris corner detector from
the regions with significant movements of human body in local video
volumes. In~\cite{lam2011pr}, motions were computed based on a
sequence of silhouette images. In~\cite{Castro14icpr}, motions were
computed on multi-view color videos, and the trajectories were
encoded by Fisher vectors for gait representation.
The model-based approaches commonly use a
priori model to match the data extracted from a
video~\cite{yam2002accv,cunado2003cviu}, and parameters of the model
are then used for gait recognition. For example,
in~\cite{cunado2003cviu}, a pendulum model is used to describe the
leg movement of the body.

Similar to~\cite{Castro14icpr}, in this paper, we also extract gait features from trajectories. However
we develop a new algorithm that is totally different from~\cite{Castro14icpr}, with availability of other sensors
and a goal to extract more accurate gait dynamics. First, we segment the walking person from the background
by using a depth sensor. This way, we can more accurately and reliably extract the human silhouette
than many human detection algorithms~\cite{felz2010pbmPami}, which only
generate rectangular bounding boxes around the person.
Second, we compute dense trajectories other than sparse
interest points and the use of dense trajectories can encode more detailed gait dynamics.



\noindent
\textbf{Depth-based methods}. With the development of
depth sensors, \textit{e.g.}, Microsoft Kinect, it is easier to
segment human body from the background and many
depth-based gait recognition methods have been proposed
recently~\cite{Sivapalan11ijcb,Munsell12eccvw,Gabel12IEMBC,Igual13eurasip,Chatto14jvcir}.
Under the assumption that body movements can be described by the
trajectories of body joints,  Munsell et al~\cite{Munsell12eccvw} proposed a
full-body motion-based method for person identification. It examines
the motion of skeletons, \textit{i.e.}, a number of joints tracked
by the Kinect, and constructs a position matrix based on the
location of the joints. All the position matrices are then dealt
with by an SVD (singular value decomposition) operation for feature
extraction. Following the idea of GEI,  Sivapalan et al~\cite{Sivapalan11ijcb}
proposed the use of GEV (gait energy volume) to represent gait dynamics with a
sequence of gait energy images, in which reasonably good recognition
accuracy can be achieved based only on the frontal depth information of
gait. However, these depth-based methods characterize
the gait dynamics only using the depth information and
neglect more detailed gait dynamics implied in the human appearance.
In~\cite{Chatto14jvcir}, PDV (pose depth volume) was used to improve GEV by
extracting accurate human silhouettes, in which color information is used
to improve the segmentation of human mask from the depth video. But PDV
does not use color information for gait representation. In~\cite{chattop2014tifs}, depth
features on body joints were obtained from Kinect depth camera, and
the GEI features were extracted from color images. The combined RGBD
features were then used for frontal gait recognition. Different
from~\cite{chattop2014tifs}, the proposed method uses color images
to compute the 2D dense trajectories, which are then combined to the depth
data to build dense 3D trajectories for extracting more detailed gait sub-dynamics.

\noindent
\textbf{Inertia-based methods}.
Early researches on inertia-based gait recognition can be found
in~\cite{Mantyjarvi05iccasp} and~\cite{Liu07icbb}.
In~\cite{Mantyjarvi05iccasp}, a portable tri-axial accelerometer
device is used, and the gait is represented by the correlation of
acceleration curves and the distribution of acceleration signals in
the frequency domain. In~\cite{Liu07icbb}, a template matching
strategy is used for gait based person identification, in which the
acceleration signals are divided by gait cycles, and then dynamic
time warping is applied to check the similarity of two gait curves.
In~\cite{Derawi10iih} and~\cite{Gafurov10icaina}, gait cycles were
detected and cycle matching were performed to improve the accuracy
of gait recognition in the context of authentication or
identification. In recent years, smart phones equipped with
accelerometer and gyroscope have been widely used, which makes it
easier and cheaper to conduct an inertia-based gait
recognition~\cite{Kwapisz10icb,Chan11icpct,XuCMU12btas,Sun14sensors}.
In~\cite{XuCMU12btas}, a Mexican-Hat wavelet transform is applied to
the acceleration data to analyze the gait patterns, and most
discriminative features are selected based on a Fisher-ratio value.
In~\cite{Ngo14pr}, large-scale data were collected for gait
recognition, in which the accelerometer is fixed on the human body.
In~\cite{zhang2015tc}, to avoid the complications in gait-cycle
detection, signature-meaningful points (SPs) on the acceleration
curve were detected, and gait features extracted on SPs were used
for gait recognition. In~\cite{Sun14sensors}, the gyroscope is used
to rectify the orientation of the accelerometer. The acceleration
signals with orientations are calculated with autocorrelation, and
converted into the frequency domain using FFT. However, the
gyroscope commonly has a cumulative-error problem, which may lead to
an unreliable rectification and the difficulty in determining the
similarity of two gait curves. Another limitation is that the
detection accuracy of previous approaches highly relies on the very
accurate placement of the accelerometer sensor on the human body.
This strict requirement would greatly affect the usability and
flexibility of the identification system.

\begin{figure*}[t!]
    \centering
    \centerline{\includegraphics[width=0.9\linewidth]{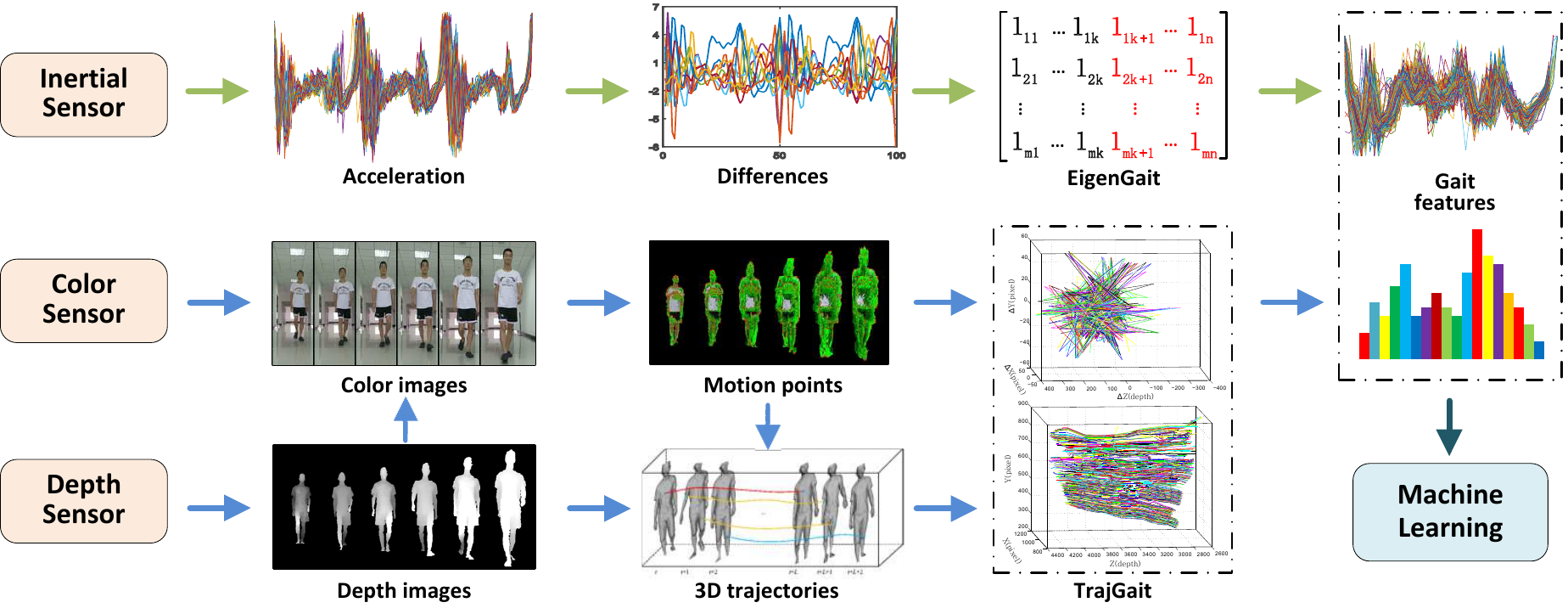}}
    \caption{{Flowchart of the proposed gait-based person identification system. }}
    \label{fig:flowchart}
\end{figure*}

\noindent
\textbf{Other action-based biometrics}.
Also related to our work is the action- or activity- based person identification~\cite{kobayashi2004action,gkalelis2009icip,iosifidis2012tifs,iosifidis2013action,lu2012mm,yan2014eccvw,yan2016neuralcomp,wu2014cvprw,kvi2015CVPR}.
Besides gait, many other actions such as jump, run and skip are also found to be capable of identifying a person. Kobayashi and Otsu~\cite{kobayashi2004action} proposed to identify persons from a sequence of motion images using an auto-correlation-based method. By incorporating more types of human actions, Gkalelis et al~\cite{gkalelis2009icip} presented a multi-modal method for person identification, and enhanced it by using a multi-camera setup to capture the human body from different viewing angles~\cite{iosifidis2012tifs}. Recently, sparse-coding-based methods were developed for human identification based on the activities captured by videos~\cite{lu2012mm,yan2014eccvw,yan2016neuralcomp}. In~\cite{lu2012mm}, a metric learning procedure was performed on the sparse-coded features to get discriminative features. In~\cite{yan2014eccvw,yan2016neuralcomp}, the discriminative power was further improved by performing a discriminative sparse projection and learning a low-dimensional subspace for feature quantization. In~\cite{wu2014cvprw}, multiple Kinects were found to improve the performance of gesture-based authentication. In~\cite{kvi2015CVPR}, a generative model was presented to describe the action instance creation process and an MAP-based classifier was used for identity inference on 3D skeletal datasets captured by Kinect.

\section{Proposed Method} \label{sec:method}

\subsection{System Overview}

Following the application scenario of person identification
shown in Fig.~\ref{fig:usage}, we
let the user walk straight along a corridor for gait feature
collection. The inertial sensors are with the user
while the color and depth sensors are placed
at the end of the corridor. In this paper, we use accelerometer in the smart phone
as inertial sensors and Microsoft Kinect as color and depth sensors. This way,
we collect the accelerometer readings and RGBD data for gait feature extraction and gait recognition.
Note that, the color (RGB) data and depth data collected by Kinect are temporally synchronized.

The flowchart of the proposed system is illustrated in Fig.~\ref{fig:flowchart}. After data
pre-processing, gait features are then extracted from the inertial data
and RGBD data by using the proposed EigenGait and TrajGait
algorithms, respectively. Finally, the gait features are combined as an input to
the machine learning component for person identification. The proposed system can be installed at the entrance of any restricted area
for person identification, such as banks, financial tower, and military base etc.

The proposed gait recognition combining multiple sensors is not fully non-invasive. The inertial
sensors move with the user and send the accelerometer data to the server. Therefore, the user
should be notified priorly and may need to show certain level of cooperation in data collection.  But from
the application perspective, most, if not all, existing person person-identification systems
for accessing a restricted area cannot be fully non-invasive -- many of them
work as a verification system where the user needs to provide
his identity to the server for verification at the entrance. For such a person-identification
system, the goal is to achieve good usability instead of full non-invasiveness.  For better usability,
a person-identification system should require
as fewer human interactions and less strict cooperations as possible.
For the proposed system, with appropriate settings and client applications in each user's
smart phone, the data collection, including sending the inertial and RGBD data, and possibly
the user's identity, to the server, and the whole process
are fully automatic without additional human interactions. In addition, as shown in the later experiments,
by combining multiple sensors, the proposed system shows higher robustness against covariate conditions.
This also improves the usability by requiring less strict cooperations from the user.

%

In the following, we first introduce the data collection and data
pre-processing, and then elaborate on the EigenGait algorithm for
inertia-based gait representation and the TrajGait algorithms for
color- and depth-based gait representation.


\subsection{Data Collection and Pre-processing}

In this paper, we use accelerometer to collect inertial data and Kinect to collect RGBD data.

\subsubsection{Acceleration data}
We utilize a tri-axial accelerometer sensor in the smart phone to
collect the acceleration data of a walking person. First, we build
an application on the Android platform. Given the APIs provided by
the Android SDK, we use the \emph{android.harware.SensorManager}
package and attached event listeners to the
\emph{Sensor.Type\_Accelerometer} to collect acceleration data. The
sensor is registered to the \emph{SensorManager.Sensor\_Delay\_Game}
and is set a sampling rate of 50Hz on each axis.

\begin{figure}[t!]
    \centering
    \subfigure[]{\label{Fig.sub2.1}\includegraphics[width=0.455\linewidth]{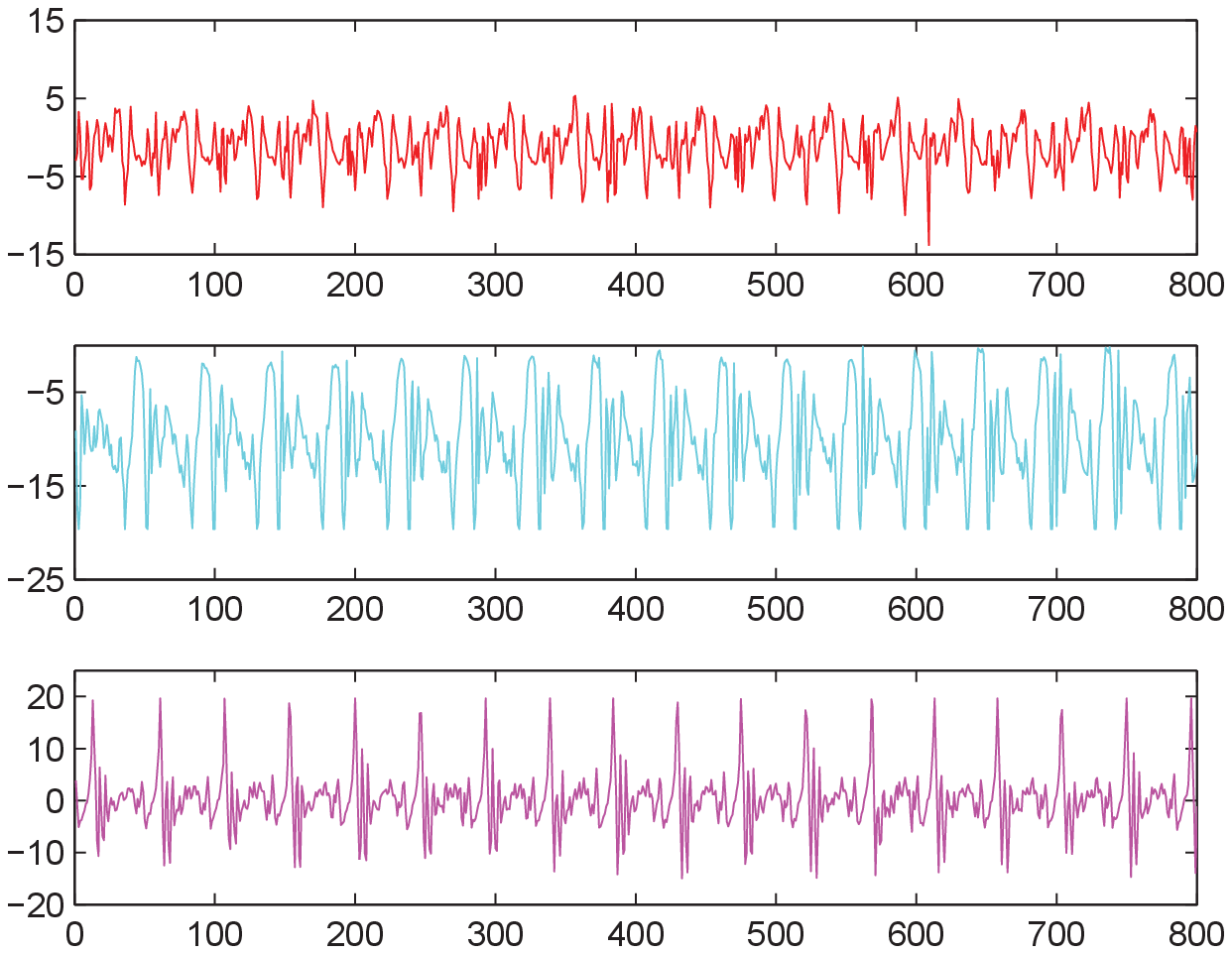}}\hspace{5mm}
    \subfigure[]{\label{Fig.sub2.2}\includegraphics[width=0.45\linewidth]{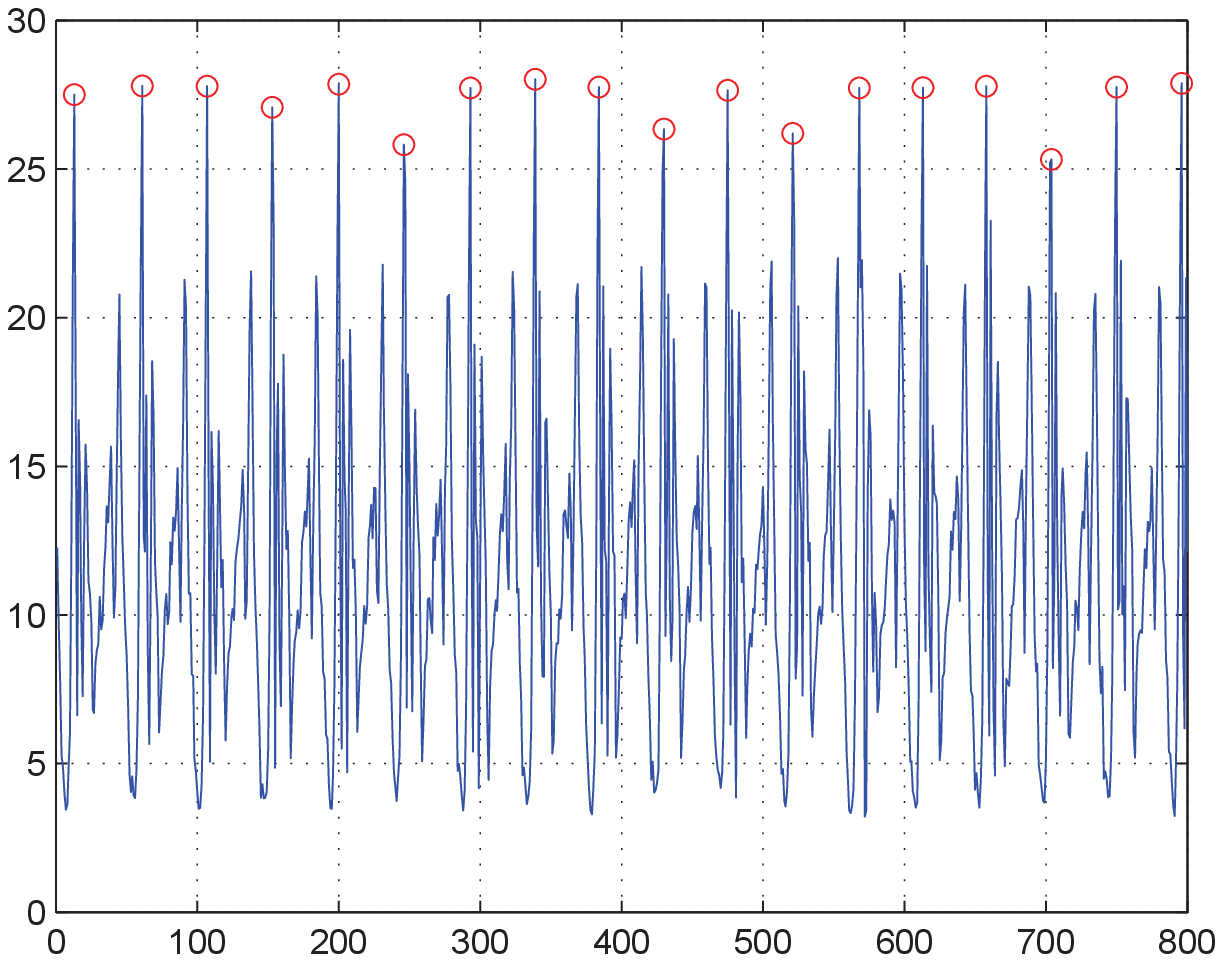}}
    \caption{{An example of the acceleration data and its partitions. (a) The acceleration values $Acc_x$, $Acc_y$ and
    $Acc_z$ on the X, Y, and Z axes, respectively. (b) The compound acceleration
    values $Acc_c$, and the partitioning points (as marked by red circles). }}
    \label{fig:acc-xyzc}
\end{figure}

Considering the usability, in data collect we simply ask the user to put the smart phone,
installed with our application, in his/her pocket with any orientation.
Each user is required to walk in his/her normal pace and fast
pace. Since the accelerometer is placed in the pocket with a random
orientation, which varies over time during the walking,
the acceleration values on each axis are collected in a time-varying direction.
Therefore, the acceleration values along each axis are actually not comparable
from time to time. To address this issue, we fuse the acceleration values on
all three axes into one compound one. Let $Acc_x$, $Acc_y$ and
$Acc_z$ be the acceleration values on the X, Y, and Z axes,
respectively, we compute the compound acceleration value $Acc_c$ by
{ $Acc_c$=$\sqrt{Acc_x^2+Acc_y^2+Acc_z^2}$}, which is more robust
against the pose change of the accelerometer over time.

Figure~\ref{Fig.sub2.1} shows an acceleration data sample on the X,
Y and Z axes collected by a smart phone -- the periodical property of the
acceleration data reflects the walking pace of the user.
Figure~\ref{Fig.sub2.2} shows the compound acceleration curve, which
has been partitioned at local maximum. Specifically, we sequentially consider
a point as the partitioning point if it satisfies three conditions: 1) it is a local
maximum (peak) along the curve, 2) its distance to the previous
partitioning point is no less than 700ms, and 3) its value is greater than 4$m$/$s^2$.

\begin{figure}[t!]
    \centering
    \subfigure[]{\label{Fig.sub4.1}\includegraphics[width=0.46\linewidth]{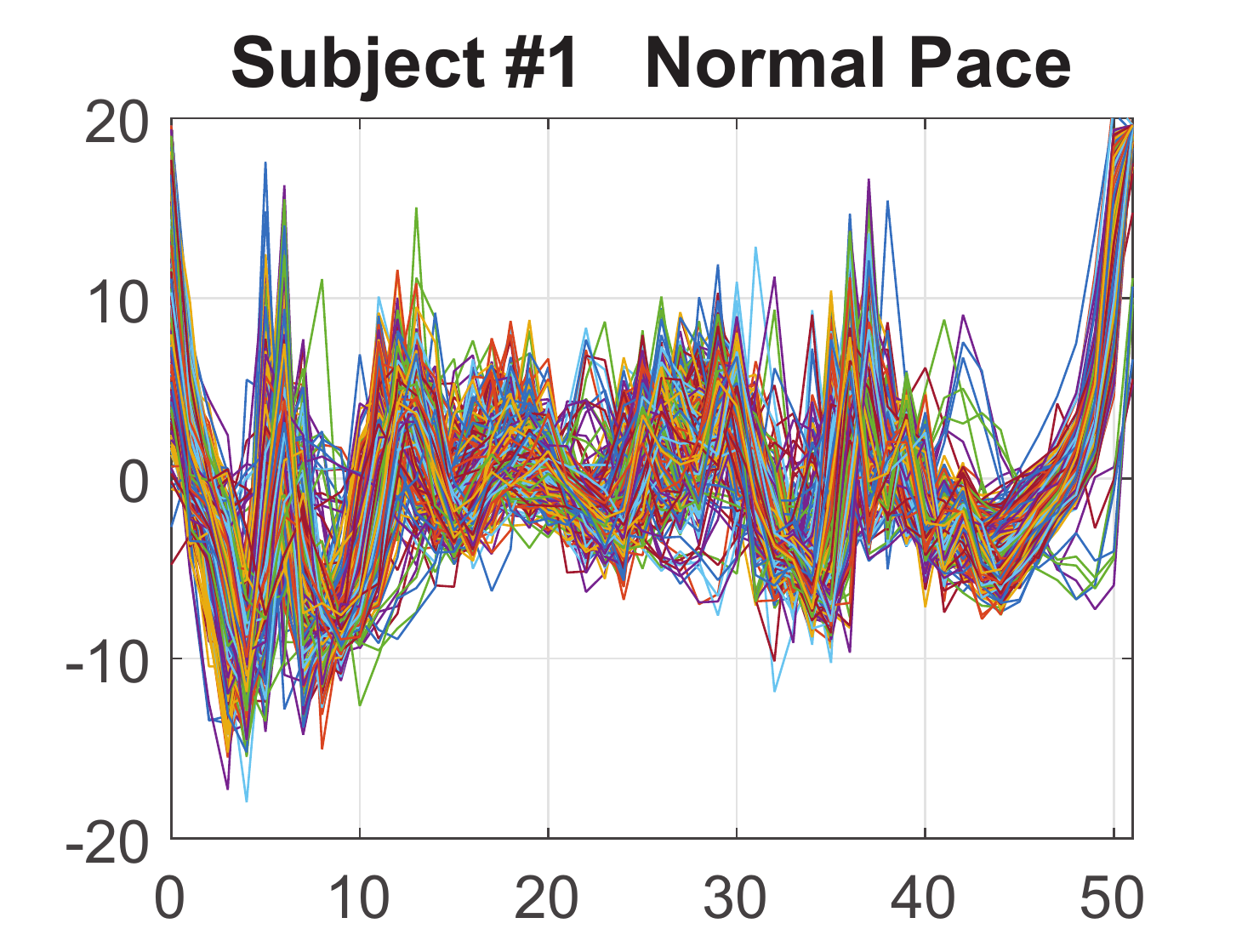}}
    \subfigure[]{\label{Fig.sub4.2}\includegraphics[width=0.46\linewidth]{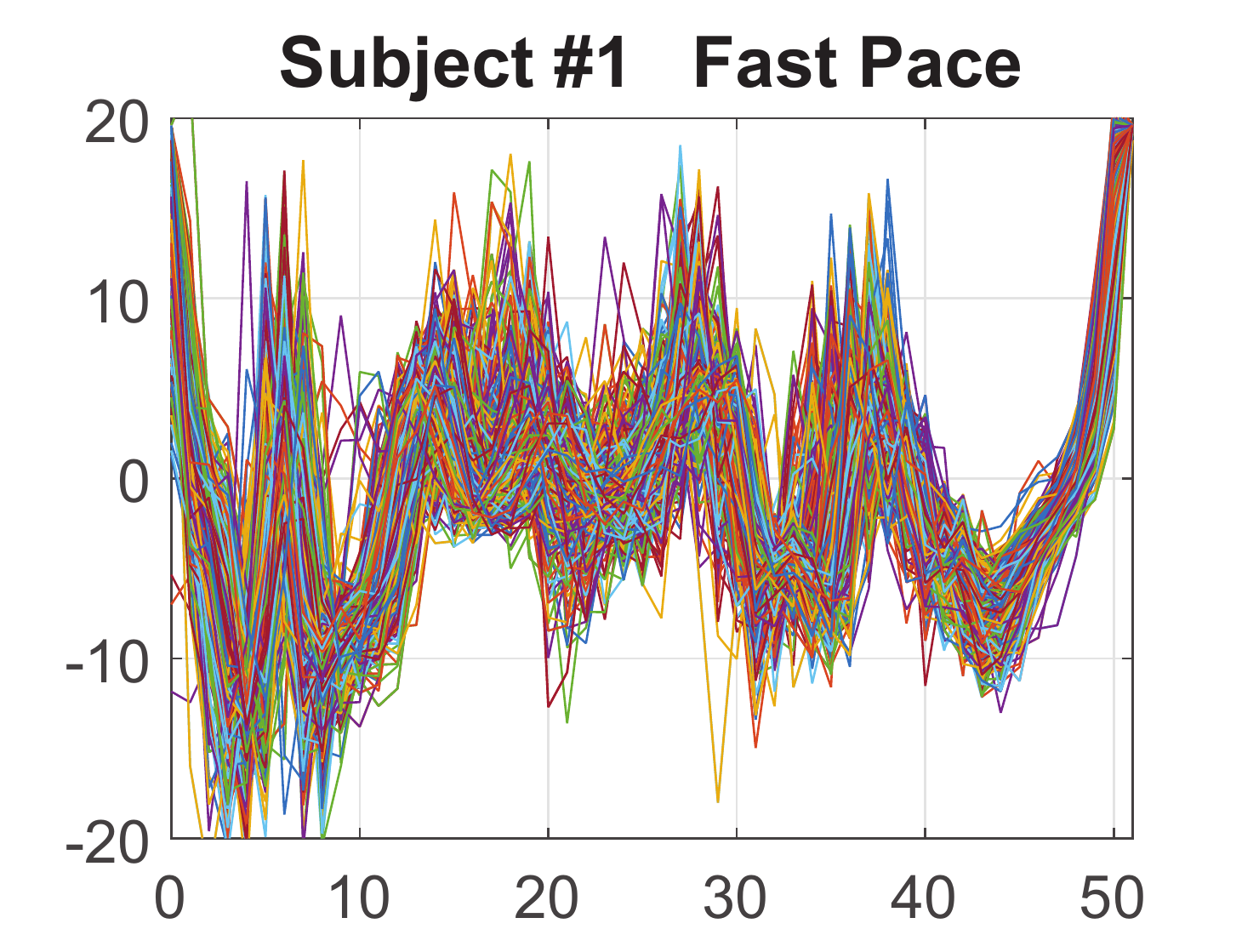}}\\
    \subfigure[]{\label{Fig.sub4.3}\includegraphics[width=0.46\linewidth]{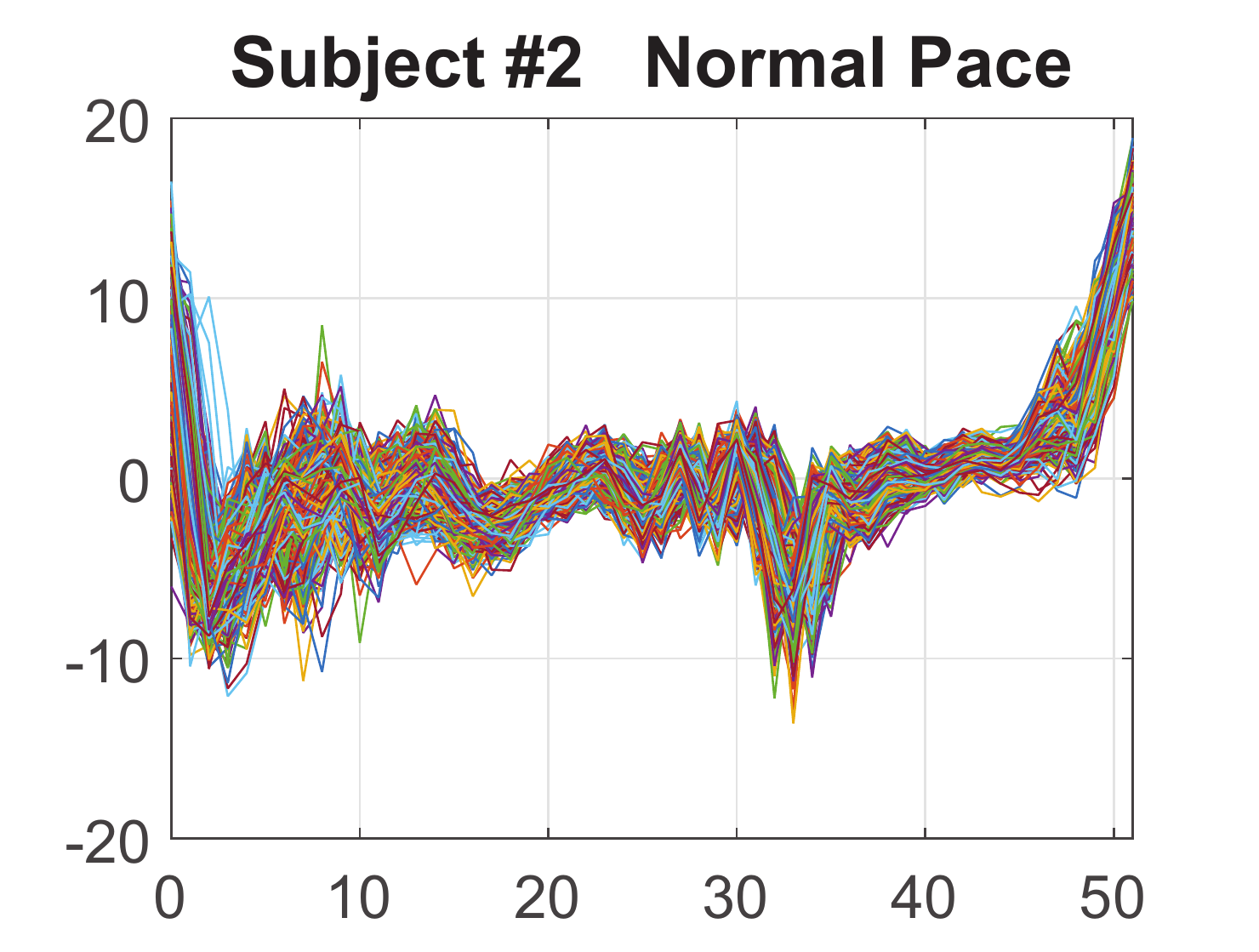}}
    \subfigure[]{\label{Fig.sub4.4}\includegraphics[width=0.46\linewidth]{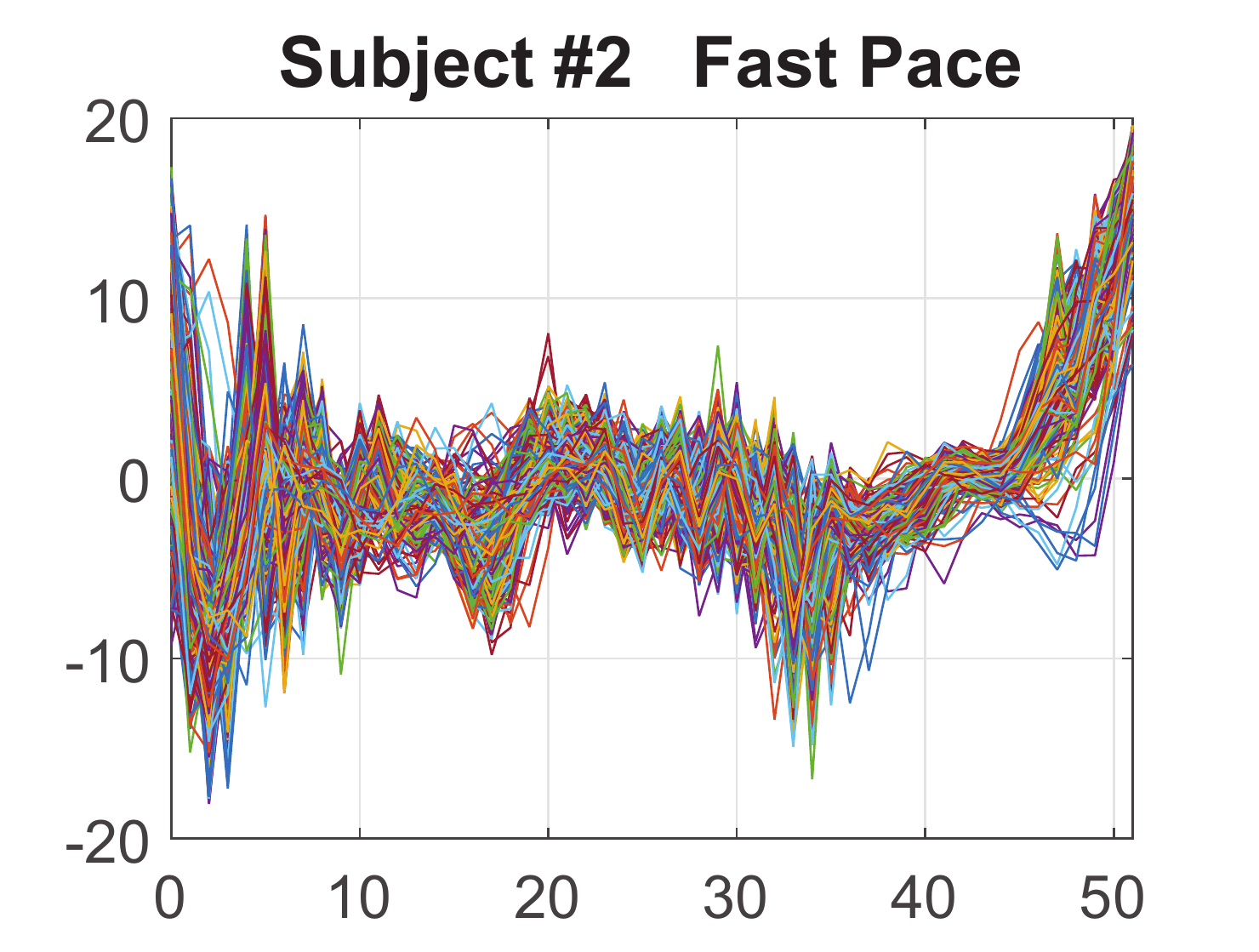}}
    \caption{{Comparison of acceleration curves collected by different subjects and on different walking paces.
    (a)\&(b) Acceleration curves of user \#1 in normal pace and fast pace, respectively.
    (c)\&(d) Acceleration curves of user \#2 in normal pace and fast pace, respectively. }}
    \label{fig:normal-fast}
\end{figure}

Each segment of the partitioning acceleration curve corresponds to one step in the walking.
Note that, in our study, one step denotes a full step cycle consisting of a
left-foot move and a right-foot move. Figures~\ref{fig:normal-fast}
(a) and (b) show 100 one-step acceleration-curve segments of an
user, under normal pace and fast pace, respectively, and Figures~\ref{fig:normal-fast} (c) and (d)
show those of another user. We can see that, although the
acceleration curves vary a lot between different users, the
acceleration curves of the same user share similar shapes, even under different paces.

\subsubsection{Color and depth data}

\begin{figure}[t!]
    \centering
    \centerline{\includegraphics[width=0.95\linewidth]{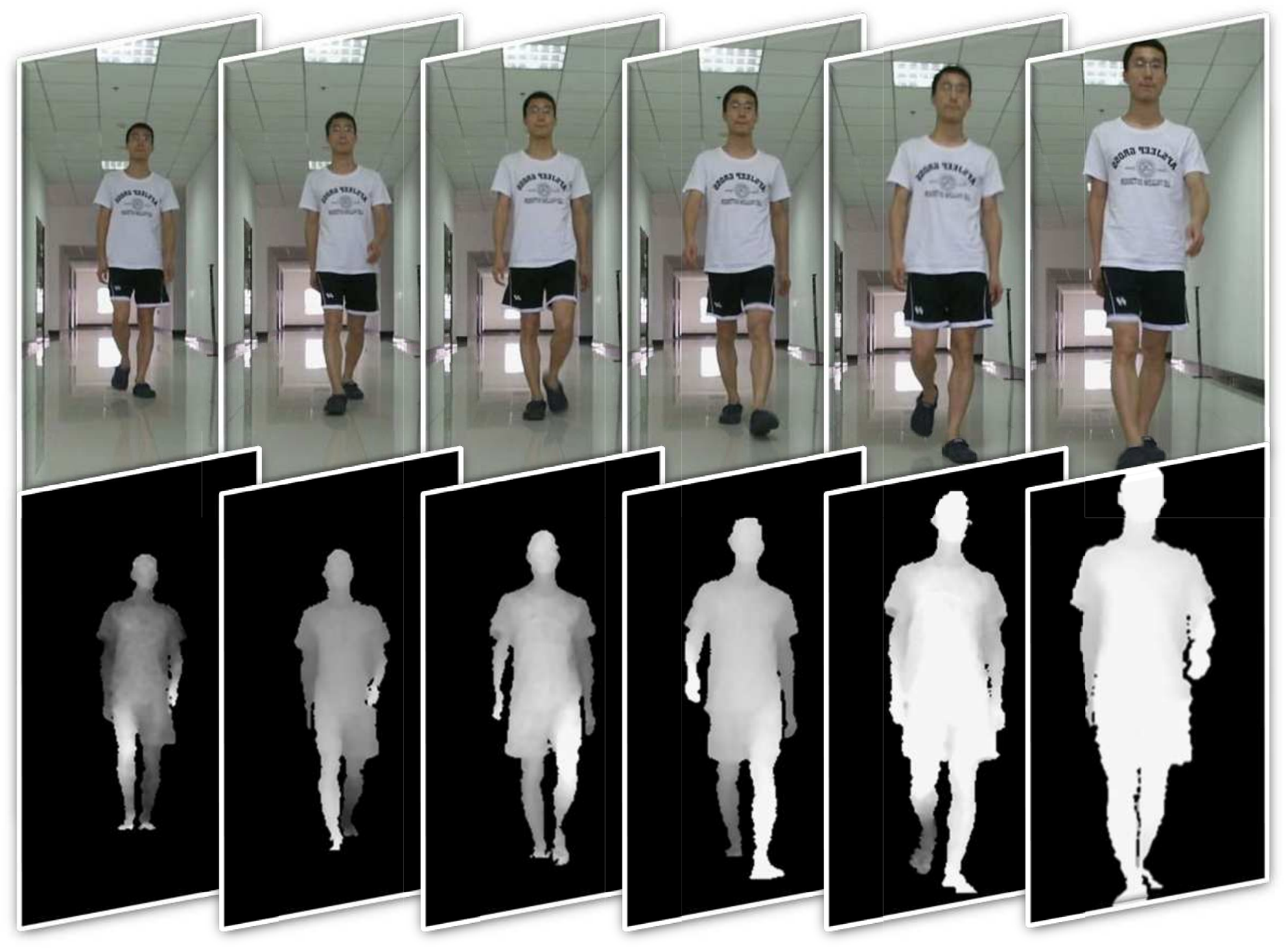}}
    \caption{{Color and depth data collected by Kinect. Top row: a sequence of
    RGB images show a person walking towards the sensors.
    Bottom row: the corresponding depth images. Note that, to give a better display,
    we crop the images by only showing the region around the person.}}
    \label{fig:kinect}
\end{figure}

\begin{figure*}[t!]
    \centering
    \centerline{\includegraphics[width=0.98\linewidth]{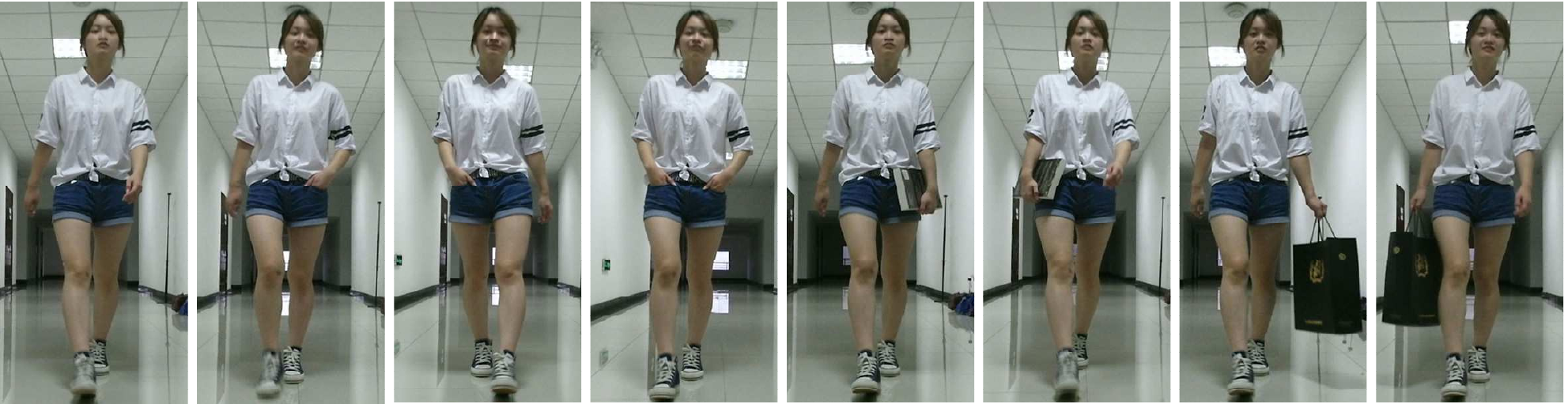}}
    \caption{{Data collection under eight different hard-covariate conditions. }}
    \label{fig:model8}
\end{figure*}

A Kinect 2.0 assisted with Kinect SDK v2.0{\footnote{
http://www.microsoft.com/en-us/kinectforwindows/develop/downloads-docs.aspx}}
is applied for color and depth data collection. The Kinect is placed
about 0.5m up from the ground. The RGB video stream is in 24-bit
true color format with a resolution of 1280$\times$1024 pixels. The
depth video stream is in VGA resolution of 640$\times$480 pixels,
with 13-bit depth value. The depth sensor has a practical ranging
limit of 1.2-3.5m distance when using the Kinect SDK. The sampling
rate is 15 fps. Figure~\ref{fig:kinect} shows a sequence of color
images and depth images collected by the Kinect. The
depth images shown in Fig.~\ref{fig:kinect} have been normalized
since a single VGA channel has only 8 bits to represent a pixel. For the
computation in all the experiments, the  original 13-bit depth value is used, which provides a
high precision to describe the motion in the depth channel.

\subsubsection{Three datasets}\label{sec:dataset}

Using the sensor settings as described above, we collect three datasets consisting
of both RGBD data and accelerometer readings. We use these data
for evaluating the performance of the
proposed method, as well as the comparison methods, in the later experiments.

\begin{table*}[htbp!]
\centering \caption{Description of the three collected datasets.}\label{tb:data}
\vspace{-2mm}
\begin{tabular}{|c|c|c|c|c|c|c|}
\hline Dataset name & Number of subjects & \tabincell{c}{Acceleration \\ data} & RGBD data & Sub-datasets & Walking pace & Other information\\
\hline \hline
{Dataset \#1} & \tabincell{c}{10\\ Male/Female: 7/3} & \tabincell{c}{1,000 groups \\ Normal/Fast: 1:1 } & \tabincell{c}{1,000 groups \\ Normal/Fast: 1:1 } & \tabincell{c}{1-step: 5,000 samples, \\ 2-steps: 5,000 samples, \\ 3-steps: 5,000 samples, \\ 4-steps: 5,000 samples, \\ 5-steps: 5,000 samples, \\ for acceleration data.} & Normal, Fast & \tabincell{c}{Acceleration data \\ and RGBD data are \\ collected independently.}\\
\hline
{Dataset \#2} & \tabincell{c}{50\\ Male/Female: 39/11} & 500 samples & 500 samples & 2-steps: 500 samples & Normal & \tabincell{c}{Acceleration data \\ and RGBD data are \\ collected at the same time.}\\
\hline
{Dataset \#3} & \tabincell{c}{50\\ Male/Female: 39/11} & \tabincell{c}{2,400 samples \\ Normal/Fast: 1:1 } & \tabincell{c}{2,400 samples \\ Normal/Fast: 1:1 } & 2-steps: 2,400 samples & Normal, Fast & \tabincell{c}{Acceleration data and \\ RGBD data are collected \\  at the same time, under \\ 8 covariate conditions.}\\
\hline
\end{tabular}
\vspace{2mm}
\end{table*}

\begin{enumerate}[$\vcenter{\hbox{\tiny$\bullet$}}$]
\item \textbf{Dataset \#1}. This dataset is collected on 10 subjects,
containing 1,000 groups of acceleration data and 1000 groups of RGBD
data -- 100 groups of acceleration data and 100
groups of RGBD data are collected for each subject, with half in normal pace, and half in fast
pace. The acceleration data and RGBD data are collected separately.
In collecting acceleration data, each subject is required to walk
along a hallway, with a length of about 60 feet. \emph{A group of
acceleration data} is defined as the sequence of acceleration values resulting from
the entire walk from one end of the hallway to the other end.
We partition the acceleration data into steps as illustrated in
Fig.~\ref{fig:acc-xyzc}(b). For all the one-step acceleration data, we
temporally interpolate them into a data sequence of length 50.
Based on the temporally partitioning, we create 5 sub-datasets,
containing one-, two-, three-, four- and five-step long data samples,
respectively. In RGBD data collection, each subject is required to
walk towards the Kinect 100 times, from about 5m away to 1m away to
the Kinect. The  sequences of frontal color and depth images of the subjects are
captured.  \emph{A group of RGBD data} is defined as the sequence of RGBD
images resulting from one full walk toward the Kinect. \vspace{2mm}

\item \textbf{Dataset \#2}. This dataset contains 500 data samples of
50 subjects, with 10 data samples for each subject. Each \emph{data sample}
consists of a sequence of acceleration data and a sequence of RGBD data, which are
collected simultaneously for one full walk of a user. For each RGBD video, a frame
is preserved only if the present person is recognized with all the body joints by the Kinect
SDK. Each acceleration data covers about 2 steps or more. We
uniformly partition each acceleration data and generate a two-step
data sample.
\vspace{2mm}

\item \textbf{Dataset \#3}. This dataset contains 2,400 data samples of
50 subjects, with 48 data samples for each subject. These data
are collected under different covariate conditions. In particular,
in collecting Dataset \#3,  each subject is required to walk under eight
different conditions, i.e., natural walking, left hand in pocket,
right hand in pocket, both hands in pocket, left hand holding a
book, right hand holding a book, left hand with loadings, and right
hand with loadings, as shown in Fig.~\ref{fig:model8}. For each
subject, 6 data samples are collected under each condition, with 3
in fast pace and 3 in normal pace. Acceleration data
and RGBD data are collected simultaneously in each data sample. The information
of the above three datasets is summarized in
Table~\ref{tb:data}.\vspace{2mm}

\end{enumerate}

\begin{figure*}[t]
    \centering
    \centerline{\includegraphics[width=0.99\linewidth]{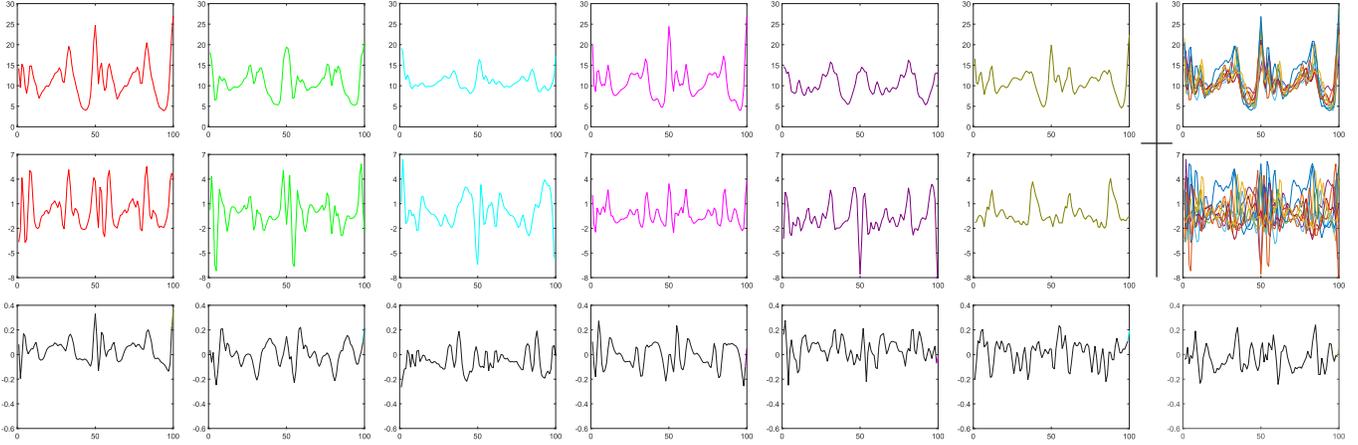}} 
    \caption{{An example for EigenGaits computation. Top row: the left six figures show the average gait curve ($\hat{\mathcal{S}_i}$) of six subjects in Dataset \#1, the last one gives an overlay view of the total ten curves.
    Middle row: the left six show the gait-curve differences (${\mathcal{O}_i}$) of the six subjects in the top row, respectively, and the last one gives an overlay view of ten gait-curve differences.
    Bottom row: from left to right, the top seven eigenvectors (EigenGaits, $\mathcal{U}$)  computed on Dataset \#1.}}
    \label{fig:eigengait}
\end{figure*}

\subsection{EigenGait: eigenspace feature extraction for gait representation}

A sequence of (compound) acceleration values resulting from a walk can be plotted into a 2D curve,
as illustrated in Fig.~\ref{fig:normal-fast} and we call it a {\it gait curve} in this paper.
Inspired by the Eigenface algorithm~\cite{turk1991eigenfaces} used for
image-based face recognition, we propose an EigenGait algorithm for
gait recognition based on gait curves.

Let $\mathcal{A}$ = $\{\mathcal{S}_{i}|i$ = $1, 2, ..., N\}$ be a
set of gait curves of $N$ subjects, $\mathcal{S}_{i}$ denotes the
gait curves collected for the $i$th subject. Treating a gait curve as a
vector, we can compute an average gait curve for the $i$th
subject as
\begin{equation}\label{eq:eigen2}
\hat{\mathcal{S}_{i}} =
\frac{1}{M_i}\sum_{j=1}^{M_i}\mathcal{S}_{i}^{(j)},
\end{equation}
where $M_i$ is the total number of gait curves collected for the
$i$th subject, and $\mathcal{S}_{i}^{(j)}$ is the $j$th gait curve
of the $i$th subject. Further, the overall average gait curve over all the
$N$ subjects can be calculated by
\begin{equation}\label{eq:eigen3}
\hat{\mathcal{S}} = \frac{1}{N}\sum_{i=1}^N\hat{\mathcal{S}_{i}}.
\end{equation}
Then, a gait-curve difference can be calculated by
\begin{equation}\label{eq:eigen4}
{\mathcal{O}_i} = \hat{\mathcal{S}_{i}}-\hat{\mathcal{S}}.
\end{equation}
To better illustrate the meaning of $\mathcal{O}_i$, we compute them
on real data. Without loss of generality, let us consider
the 2-step acceleration data collected in Dataset \#1. In Fig.~\ref{fig:eigengait},
the last figure in the middle row shows the gait-curve differences
of ten subjects in Dataset \#1. It can be seen from
Fig.~\ref{fig:eigengait} that, the gait-curve differences also
preserve the periodic property of the original gait curve, as shown
in the top row of Fig.~\ref{fig:eigengait}, and different subjects
have different gait-curve differences.

Then the covariance matrix can be calculated by
\begin{equation}\label{eq:eigen-cov}
{\mathcal{C}} =
\frac{1}{N}\sum_{i=1}^N{\mathcal{O}_{i}}{\mathcal{O}_{i}^{\top}}.
\end{equation}
We can perform an eigen-decomposition as
\begin{equation}\label{eq:eigen-pca}
{(\mathcal{\lambda},\mathcal{U})} = \mathbf{Eigen}({\mathcal{C}}),
\end{equation}
where $\mathcal{\lambda}$ denotes the eigenvalues, and $\mathcal{U}$
denotes the corresponding eigenvectors. Suppose the eigenvalues in
$\mathcal{\lambda}$ have been sorted in descending order, we select
the first $r$ elements that fulfill $\sum_{i=1}^{r}\lambda_i
\geq 0.85\cdot\sum \mathcal{\lambda}$, and hence get $r$
corresponding eigenvectors $\{u_1, u_2, ..., u_{r}\}$. In the bottom
row of Fig.~\ref{fig:eigengait}, the seven curves show the top seven
eigenvectors of the two-step sub-dataset in Dataset \#1. It can be
seen from Fig.~\ref{fig:eigengait} that, more distinctiveness can be
observed in the gait-curve differences than in the original gait
curves. We can also see that, these eigenvectors preserve the shape
appearance of some of the original gait curves, as shown in the top
row of Fig.~\ref{fig:eigengait} and we call
them {\it EigenGaits} in this paper. When a new gait curve $s$ comes, we can
project it into the eigenspace defined by the $r$ eigenvectors as
\begin{equation}\label{eq:eigen-prj}
{\mathcal{\omega}_i} = {u_{i}^{\top}}(s-\hat\mathcal{S}),\ i=1,2,
..., r,
\end{equation}
and obtain an EigenGait feature vector $(\mathcal{\omega}_1,
\mathcal{\omega}_2, ..., \mathcal{\omega}_{r})$. As the acceleration
data reflects the whole body motion in the walking, the extracted
EigenGait features can capture the general gait dynamics.

\subsection{TrajGait: dense 3D trajectories based gait representation}
The gait data captured by color sensor and depth sensor can be
represented by a sequence of color images and depth images, respectively.
These images provide useful information to describe the details of body
movements, e.g., the movement of each body part.
We combine the color and depth data and develop
a {\it TrajGait} algorithm for extracting 3D dense trajectories
and describing more detailed gait sub-dynamics.

%


\begin{figure}[t!]
    \centering
    \centerline{\includegraphics[width=0.99\linewidth]{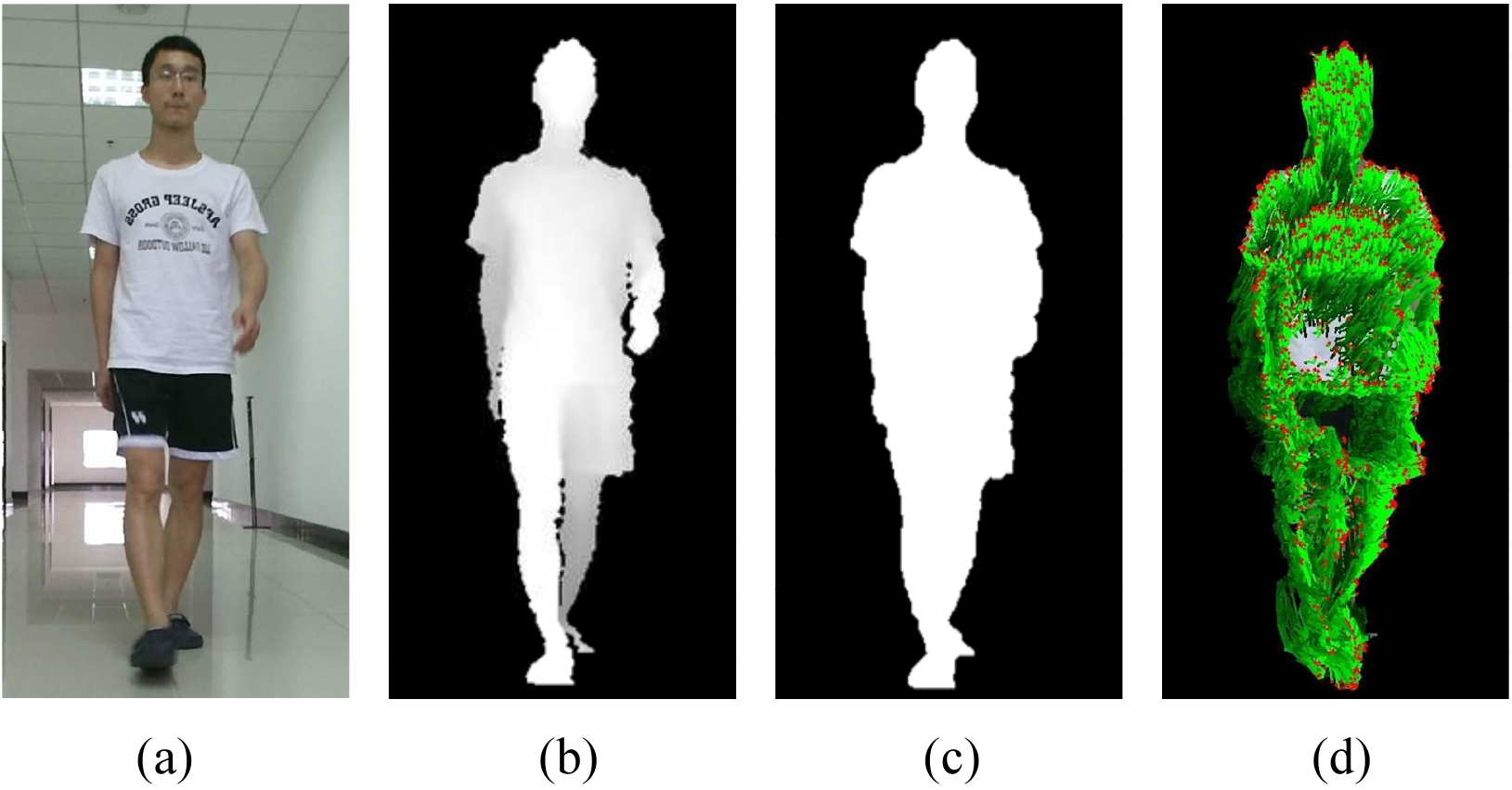}}
    \caption{{An example of dense trajectory points extraction. (a) A color image, (b)
    the corresponding depth image, (c) the segmented mask image, and (d) 2D dense
    trajectories within the mask, where the red dots indicate the point positions
    in the current frame. Note that, the image mask has been fine-tuned with
    image operations, including hole filling, noise removal, morphological operation. }}
    \label{fig:kinectMask}
\end{figure}

\begin{figure*}[t!]
    \centering
    \centerline{\includegraphics[width=0.6\linewidth]{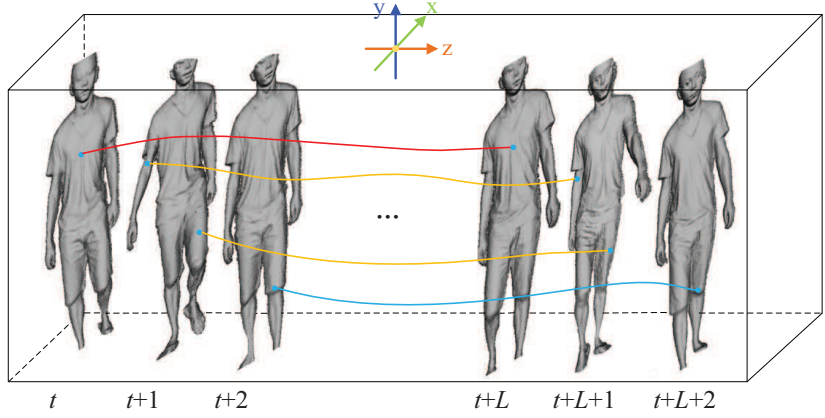}}
    \caption{{Illustration of the trajectories in 3D space. }}
    \label{fig:traj-3d}
\end{figure*}

TrajGait algorithm is summarized in Algorithm 1,
which contains the following four key operations:

\begin{enumerate}[$\vcenter{\hbox{\tiny$\bullet$}}$]

\item \textbf{computMotion} One each RGB
color frame, we compute the dense optical flow by the algorithm
proposed by F$\ddot{a}$rneback~\cite{farneback2003two}{\footnote{ It
is implemented and released by OpenCV 2.4.8 or above.}}. This
algorithm makes a good compromise between accuracy and speed.
\vspace{2mm}

\item \textbf{segmentMask} To focus on the walking person,
we segment the person from the background, and take
it as a mask in later operations. Since the Kinect SDK has provided
functions for efficient human detection and joints
tracking~\cite{Shotton13pami}, we apply these functions to extract a
raw human mask in each frame, and then apply some image processing
techniques, including hole filling and noise removal, to get the final mask.
Figure~\ref{fig:kinectMask}(c)
displays a human mask segmented from the depth image in
Fig.~\ref{fig:kinectMask}(b).
Note that, while segmenting persons
from a confusing background can be very challenging on RGBD data, it is not a serious
issue in the proposed application scenario of person identification -- the
environment is highly controlled (e.g., a hallway) and the sensors are well set, without any other moving
objects around. In this paper, the following steps are taken to obtain the human mask: (i) produce human-oriented depth image using the body-segmentation function provided by Kinect SDK (i.e., IBodyIndexFrame::AccessUnderlyingBuffer), (ii) resize the depth image to the size of the color image and interpolate the resized image using bi-cubic interpolation, (iii) binarize the depth image with a threshold $t$=113, and (iv) fill the holes and remove segments that are smaller than 1,000 pixels.


\vspace{2mm}

\begin{algorithm}[htbp]\small{
\caption{\ ~TrajGait algorithm}\label{al:gait-f}
\begin{algorithmic}[1]
    \Procedure{TrajGait}{}
       \State \textbf{input:}
       \State \quad $V_1, V_2, ..., V_N$:~ RGB data collected for $N$ subjects,
       \State \quad $D_1, D_2, ..., D_N$:~the corresponding depth data,
       \State \quad $X_1, X_2, ..., X_N$:~the number of data samples in each set,
       \State \quad $\mathcal{K}$:~the number of centers in the K-means clustering,
       \State \quad $L$:~the number of frames in a trajectory,
       \State \textbf{output:}
       \State \quad $\{H_i|i=1,2, ..., \mathcal{X}\}$:~ feature histograms for all RGBD
       \State \qquad  \ \ \ videos, where
       $\mathcal{X}=\sum_1^NX_i$.

      \Statex \vspace{-1.9mm}
      \State \% {Calculate the trajectories of all RGBD data}:
       \For{($i$=1 to $N$)}
       \For{($j$=1 to $X_i$)}
            \State \% {Compute the motion on the color video $V_i^{(j)}$:}
            \State $\mathcal{M}_i^{(j)} \gets \textbf{computMotion}(V_i^{(j)})$;
            \State \% {Segment foreground (human) from the depth video:}
            \State Mask$_i^{(j)} \gets \textbf{segmentMask}(D_i^{(j)})$;
            \State \% {Calculate 3D trajectories in the RGBD channel:}
            \State $\mathcal{T}_i^{(j)} \gets \textbf{calcTrajectories}(\mathcal{M}_i^{(j)}, D_i^{(j)},\ $Mask$_i^{(j)}, L)$;
            \State $\mathcal{T}_i \gets \textbf{putInto}(\mathcal{T}_i^{(j)})$;
       \EndFor
       \EndFor

       \Statex \vspace{-1.8mm}
       \State \% {Put all trajectories together}:
       \State $\mathcal{T} = \{\mathcal{T}_i|i=1,2, ..., N\}$;
       \State \% {Compute a number of $\mathcal{K}$ centers using Clustering}:
       \State $\mathcal{Y} \gets \textbf{kMeans}(\mathcal{T}, \mathcal{K})$;

       \Statex \vspace{-1.8mm}
       \State \% {Compute trajectory histogram for each RGBD data sequence}:
       \For{($i$=1 to $N$)}
       \For{($j$=1 to $X_i$)}
            \State $H_i^{(j)} \gets \textbf{histTrajectory}(\mathcal{T}_i^{(j)}, \mathcal{Y})$;
       \quad \EndFor
       \quad \EndFor
    \EndProcedure
\end{algorithmic}}
\end{algorithm}

\item \textbf{calcTrajectories}
Suppose $(x, y)$ is the coordinate of a point at a frame of the collected
color data, $(z)$ is the depth value of that point in the depth
video, then we can locate that point with a coordinate $(x_t, y_t,
z_t)$ in the RGBD space. In this way, we can treat each point in the
RGBD data as a 3D point. Figure~\ref{fig:traj-3d} illustrates the
trajectories in the 3D space. The shape of a trajectory encodes the
local motion patterns, which we use for gait
representation. Based on the 2D dense trajectories extracted
by~\cite{wang2011cvpr} in RGB channels, we can compute the corresponding 3D
trajectories.\vspace{1mm}

Let's further suppose point $P_t=(x_t, y_t, z_t)$ at frame $t$ is
tracked to frame $t$+1 at the point $P_{t+1}$, then, with a given
trajectory length $L$, we can describe its shape by a displacement
vectors,
\begin{equation}\label{eq:eigen-traj1}
\mathcal{F} = (\Delta P_t, \Delta P_{t+1}, ..., \Delta P_{t+L-1}),
\end{equation}
where $\Delta P_t = (P_{t+1} - P_t) = (x_{t+1} - x_t, y_{t+1} - y_t,
z_{t+1}-z_{t})$, and $L$ is empirically set as 15.
Since the gait may be collected in various walking
speed~\cite{kusakunniran2012tsmc}, the resulting vector has to be
normalized to reduce deviations. As the metric in the color image is
different from that in the depth image, we separately normalize them
by their sums of the magnitudes of the displacement vectors. We take
a normalized displacement vector as a 3D trajectory descriptor. An
example of 3D trajectory descriptors derived from an RGBD data sequence
is shown in Fig.~\ref{fig:traj-descriptor}.
\vspace{2mm}

\begin{figure}[t!]
    \centering
    \subfigure[]{\label{Fig.traj.1}\includegraphics[width=0.7\linewidth]{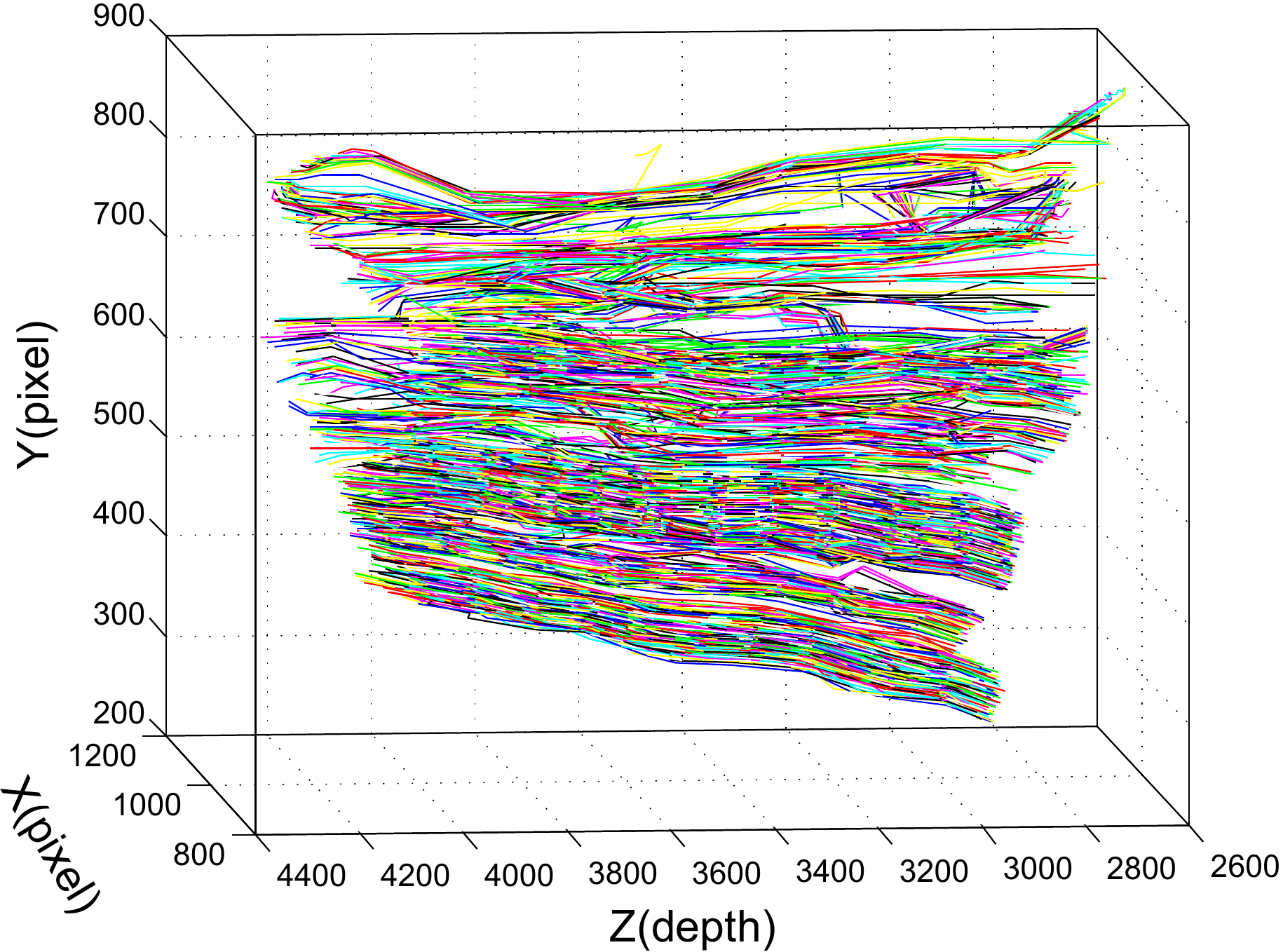}}\hspace{1mm}
    \subfigure[]{\label{Fig.traj.2}\includegraphics[width=0.7\linewidth]{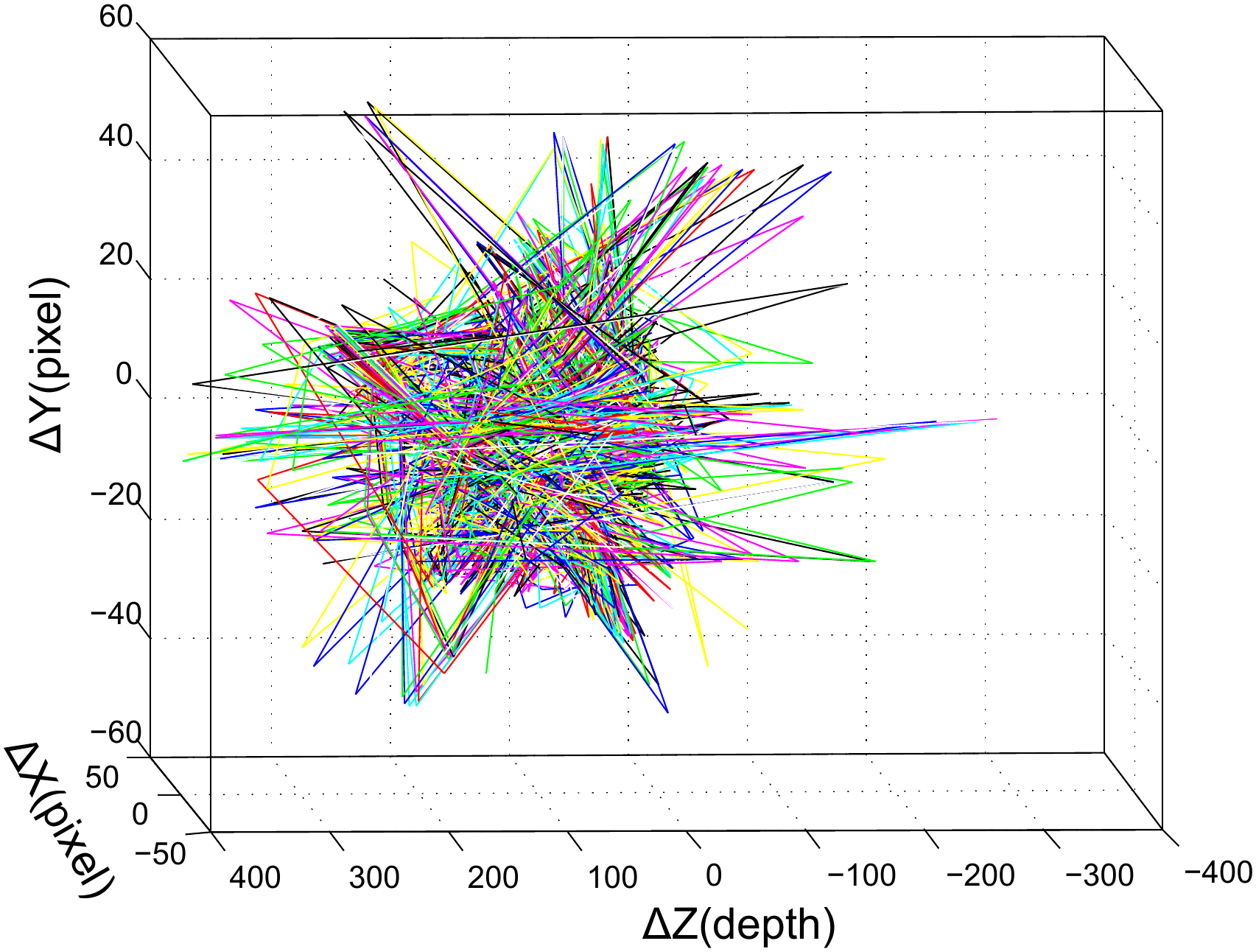}}
    \caption{{An example of the 3D trajectory descriptors. (a) 3D trajectories
    extracted from an RGBD data sequence, (b) the 3D trajectory displacements, i.e.,
    3D trajectory descriptors, computed on the trajectories in (a). }}
    \label{fig:traj-descriptor}
\end{figure}

\item \textbf{histTrajectory} We apply a bag-of-words strategy to encode
the 3D trajectory descriptors. Specifically, we generate a codebook
with a number of $K$ codes using a clustering technique. The
standard  K-means algorithm is employed here for clustering. To reduce the complexity, we
cluster a subset of 1,000,000 randomly selected training samples. To
increase the precision, we run K-means 10 times and keep the result with
the lowest K-means clustering cost. For each RGBD sequence, the extracted 3D trajectory
descriptors are quantized into a histogram by hard assignment. The
resulting trajectory histograms are then used for gait
representation.
\end{enumerate}

\subsection{Gait Recognition}

We achieve gait recognition using a supervised classifier.
We combine the gait features extracted by EigenGait and TrajGait
and feed them into a machine learning component
for training and testing. The trained model can then be used
to recognize new unseen data samples for gait
recognition and person identification. For feature combination,
we simply concatenate the EigenGait features and the TrajGait features
into one single feature vector.

In the machine learning component, a multiclass Support Vector Machine (SVM)
classifier implemented by
libSVM{\footnote{www.csie.ntu.edu.tw/~cjlin/libsvm/}} is used for
both training and testing~\cite{LinCJ11a}. A one-vs-all
classification strategy is applied. To investigate the potential
relation between classification accuracy and computation efficiency,
we try both the linear and non-linear SVMs. For the soft-margin
constant $C$ in SVM, we consistently set it 1,000 through all the
experiments.

\section{Experiments and Results} \label{sec:experiment}

\begin{figure*}[t!]
    \centering
    \subfigure[]{\label{Fig.sub10.1}\includegraphics[width=0.39\linewidth]{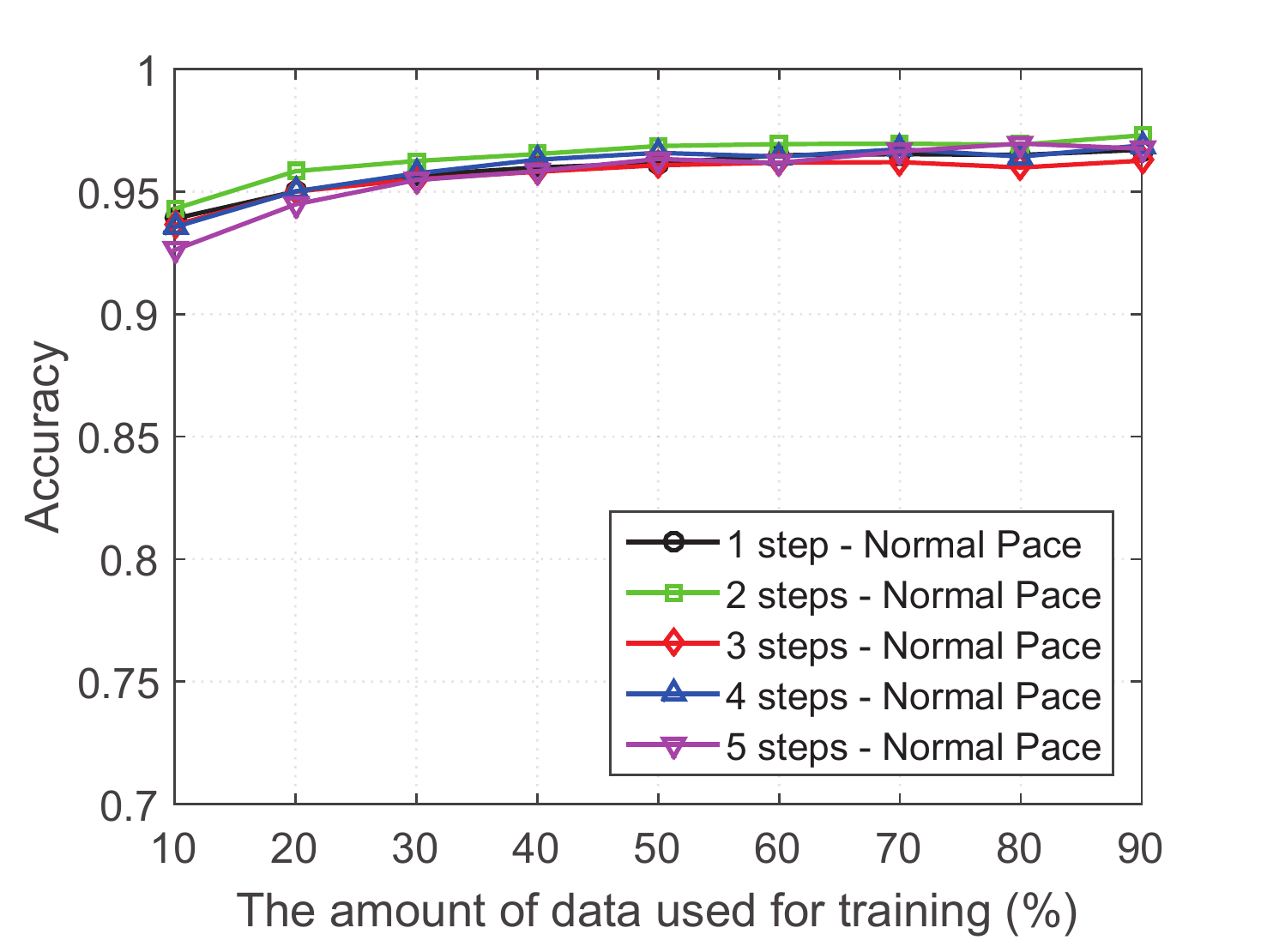}}
    \subfigure[]{\label{Fig.sub10.2}\includegraphics[width=0.39\linewidth]{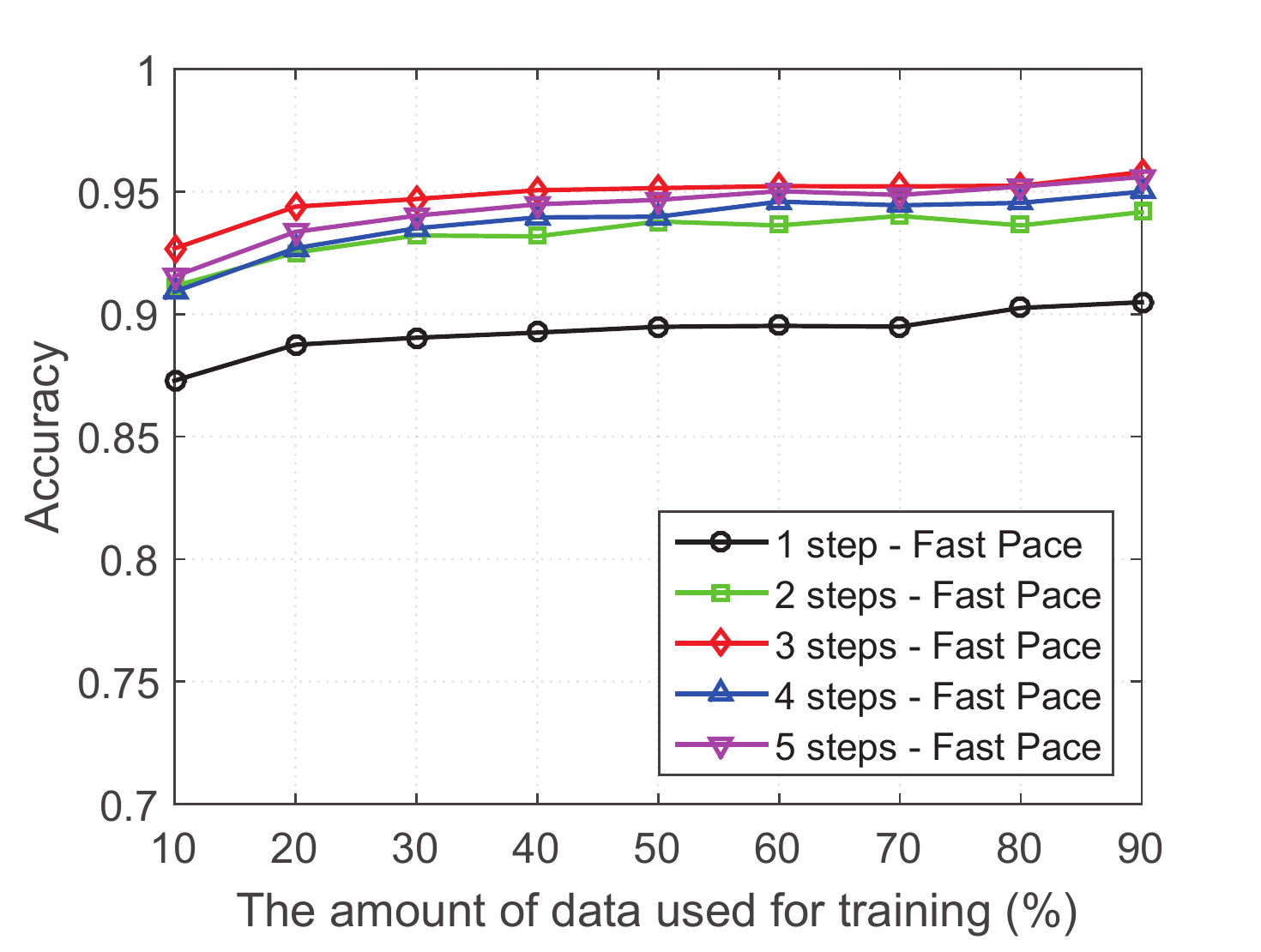}}\\ \vspace{-3mm}
    \subfigure[]{\label{Fig.sub10.3}\includegraphics[width=0.39\linewidth]{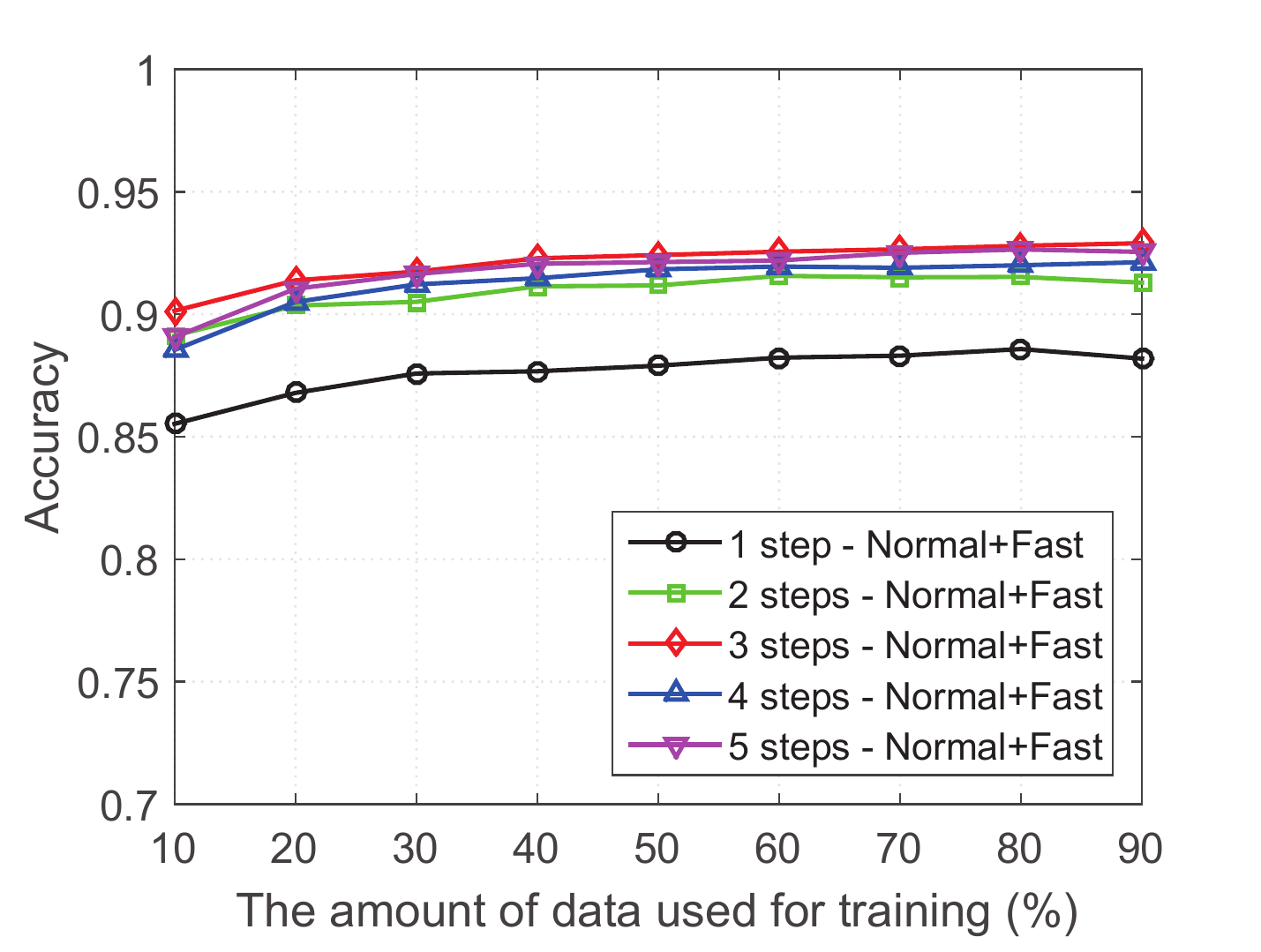}}
    \subfigure[]{\label{Fig.sub10.4}\includegraphics[width=0.39\linewidth]{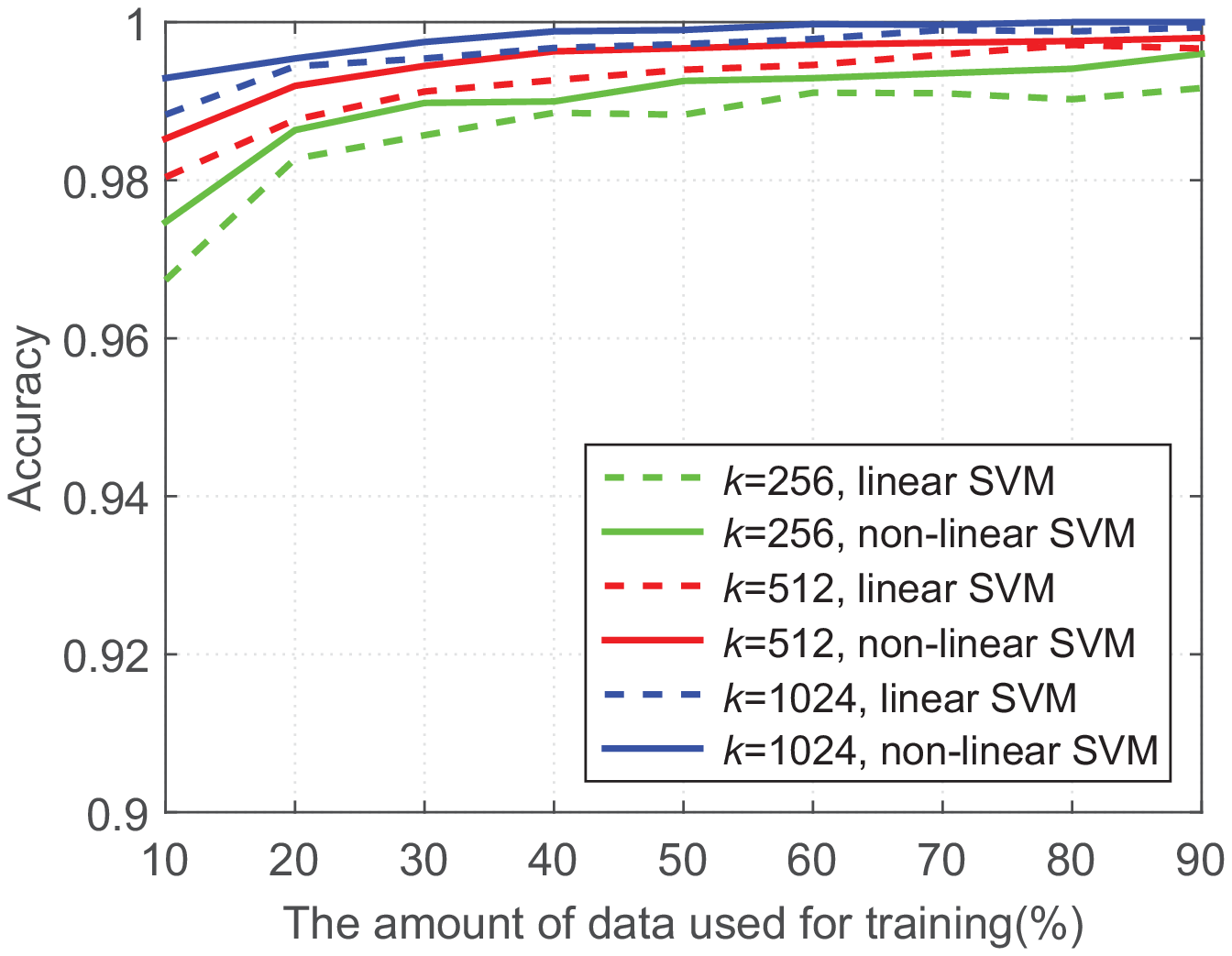}}\vspace{-2mm}
    \caption{{Performance of the EigenGait and TrajGait. (a) EigenGait on normal pace, (b)  EigenGait
    on fast pace, (c) EigenGait on normal+fast case, and (d) TrajGait on different settings. }}
    \label{fig:eigen5steps}
\end{figure*}

In this section, we use three datasets to evaluate the performance
of the proposed method, as well as the comparison methods. First, we
examine the effectiveness of the proposed EigenGait algorithm and
the TrajGait algorithm using Dataset \#1, separately. Then, we
evaluate the performance of the proposed method, i.e., the one
fusing EigenGait and TrajGait, on Dataset \#2 by comparing its
accuracy with several state-of-the-art gait recognition methods. Finally, we test the
robustness of the proposed method on Dataset \#3. In particular, we
try to answer the following questions:
\begin{enumerate}[\hspace{2mm}-]
\item How effective are the EigenGait algorithm and the TrajGait algorithm for gait
recognition? How do the parameters influence their performances?
\item What is the overall performance of the proposed method? Does it work better than the
state-of-the-art color-based methods, depth-based methods, and inertia-based methods?
\item How robust is the proposed method in handling gait data collected under hard-covariate conditions?
\end{enumerate}

In the experiments, we mainly evaluate gait recognition to address a classification problem. At the
end of the section, we will also evaluate the proposed method to solve an identification problem.
As a classification problem, we use the classification accuracy as a metric for performance evaluation.
The classification accuracy is defined as

{\small
\begin{equation}\label{eq:exp-metric}
Accuracy = \frac{\#\ Correctly\ Classified\ Samples}{\#\ Total\ Testing\ Samples}.
\end{equation}}
In the classification, each testing sample is classified by the pre-trained SVM
and receives a score vector containing $n$ score values, where $n$ is the number of
subjects in training the SVM. A score value in the score vector
indicates the likelihood of this sample to be from a specific subject.
The sample will be recognized as being from subject $i$ if the $i$th element is the maximum in the score vector.
Compared with the ground-truth subject for the test sample, we can decide whether it is correctly classified and
compute the accuracy.

\begin{table*}[t!]
\centering \caption{Performance (Accuracy) of EigenGait using linear and
non-linear SVM.}\label{tb:result}
\vspace{-2mm}
\begin{tabular}{|c|c|cccccccc|}
\hline Walking pace &Kernel &1 step &2 steps &3 steps &4 steps &5 steps &6 steps &7 steps &8 steps\\
\hline \hline
\multirow{2}{*}{Normal} &KL1 &0.9616 &0.9616 &0.9608 &0.9659 &0.9634 &0.9626 &0.9525 &0.9606\\
&KCHI2 &0.9522 &0.9510 &0.9471 &0.9520 &0.9449 &0.9454 &0.9393
&0.9354\\  \hline
\multirow{2}{*}{Fast} &KL1 &0.8948 &0.9308 &0.9515 &0.9398 &0.9467 &0.9437 &0.9387 &0.9250\\
&KCHI2 &0.8894 &0.9208 &0.9433 &0.9292 &0.9356 &0.9244 &0.9200
&0.9018\\ \hline
\multirow{2}{*}{Normal\&Fast} &KL1 &0.8790 &0.9048 &0.9242 &0.9183 &0.9213 &0.9247 &0.9094 &0.8977\\
&KCHI2 &0.8785 &0.8992 &0.9156 &0.9082 &0.9219 &0.9054 &0.8913
&0.8900\\ \hline
\end{tabular}
\vspace{2mm}
\end{table*}

\subsection{Effectiveness}

We use Dataset \#1 to evaluate the performance of the EigenGait
algorithm and the TrajGait algorithm. Dataset \#1 is collected for 10
subjects, including five sub-datasets of the acceleration data, and
one sub-dataset of  the RGBD data.

\subsubsection{EigenGait} There are five acceleration sub-datasets, i.e.,
the one-, two-, three-, four- and  five-step sub-datasets. Each sub-dataset contains 5,000
acceleration data sequences, with half in normal pace and half in fast pace. The resulting EigenGait
features are of dimension 43, 85, 128, 170 and 213 for the one-, two-, three-, four- and five-step
data, respectively. Note
that, data in the same sub-dataset have no overlaps with
each other. The EigenGait algorithm is evaluated under the normal
pace, fast pace and two paces mixed, i.e., normal+fast, and the
results are shown in Fig.~\ref{fig:eigen5steps}(a)-(c),
respectively. From Fig.~\ref{fig:eigen5steps}(a)-(c), we can see that,
EigenGait obtains good classification accuracy in all three cases,
e.g., over 0.95 in normal pace, 0.92 in fast pace, and 0.90 in
normal+fast, using 30\% data for training. Moreover, EigenGait shows
higher accuracy under normal pace than under fast pace. This is
because, a large speed variation would occur when a person walks in
a fast pace, which would increase the complexity of the gait data.
Decreased performances of EigenGait can be observed in
Fig.~\ref{fig:eigen5steps}(c), because the mixed-pace data further increase
the data complexity.

As can be seen from Fig.~\ref{fig:eigen5steps}(b) and (c), on a
dataset with large speed variations, e.g., in fast pace, or in
normal+fast, EigenGait holds lower performances on the 1-step
dataset than on two or more step dataset. This is because, a 1-step
data is less capable of representing the gait than a 2 or more
step data. Surprisingly, the EigenGait obtains comparable
performances when varying the data length from 2 to 5 steps.
Considering that a 2-step data can be easily captured and
efficiently computed as comparing to longer data, in our later
experiments, we always choose a length of 2 steps for EigenGait
features, including the experiments on Dataset \#2 and Dataset \#3.

Further, we evaluate EigenGait's performance using linear and
non-linear SVMs. Typically, the `KL1' and `KCHI2' kernels are
employed, respectively. Table~\ref{tb:result} lists the classification accuracy under
different walking paces and varied data lengths, where 50\% data are used for training. It can be seen that,
EigenGait generally shows higher performances using a linear
SVM than using a non-linear one. This is because, the EigenGait
extracts gait features in the eigenspace, which makes the feature
more linearly classifiable.

\begin{figure*}[htbp]
    \centering
    \subfigure[]{\label{Fig.sub50.1}\includegraphics[width=0.33\linewidth]{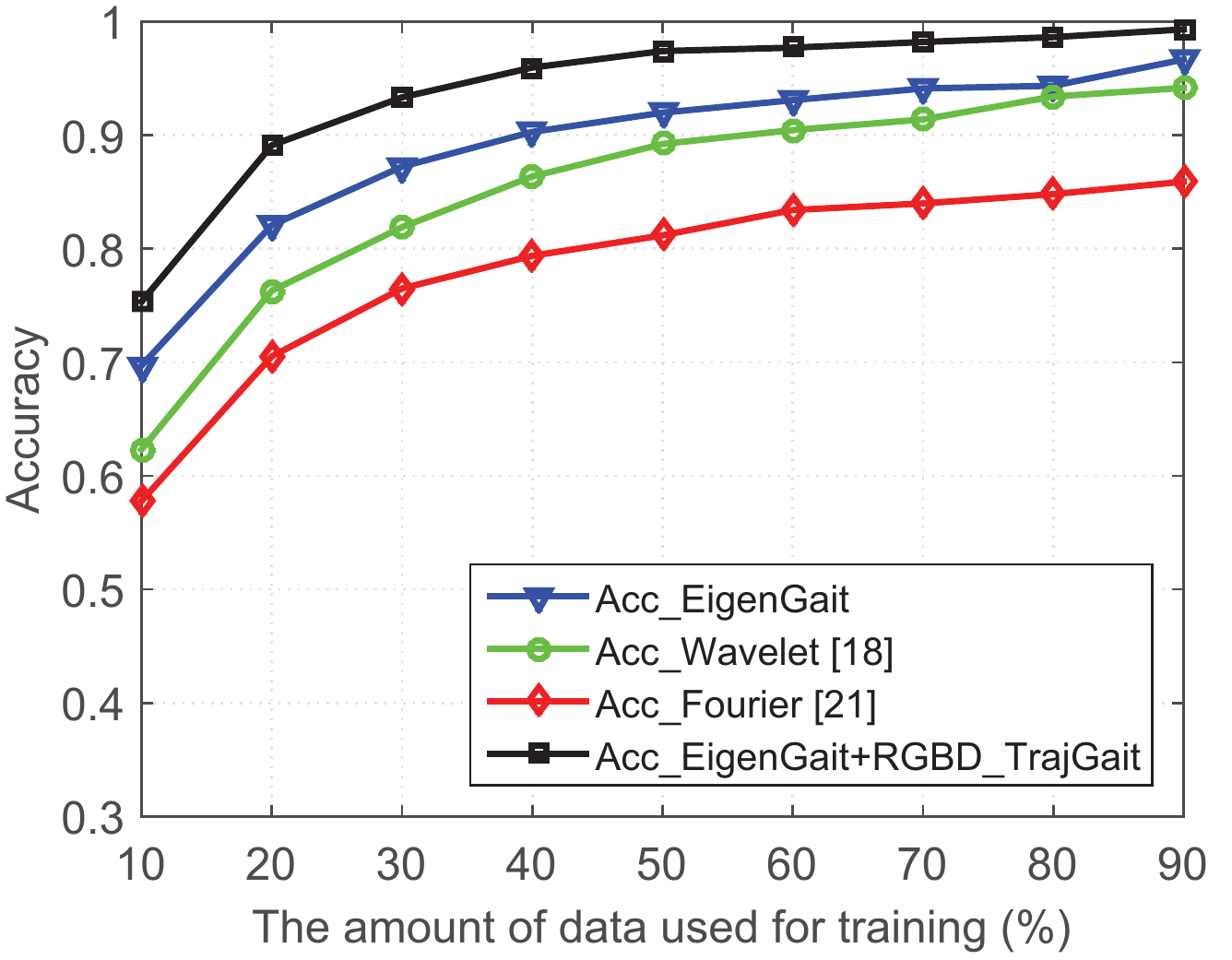}}\hspace{-2mm}
    \subfigure[]{\label{Fig.sub50.2}\includegraphics[width=0.33\linewidth]{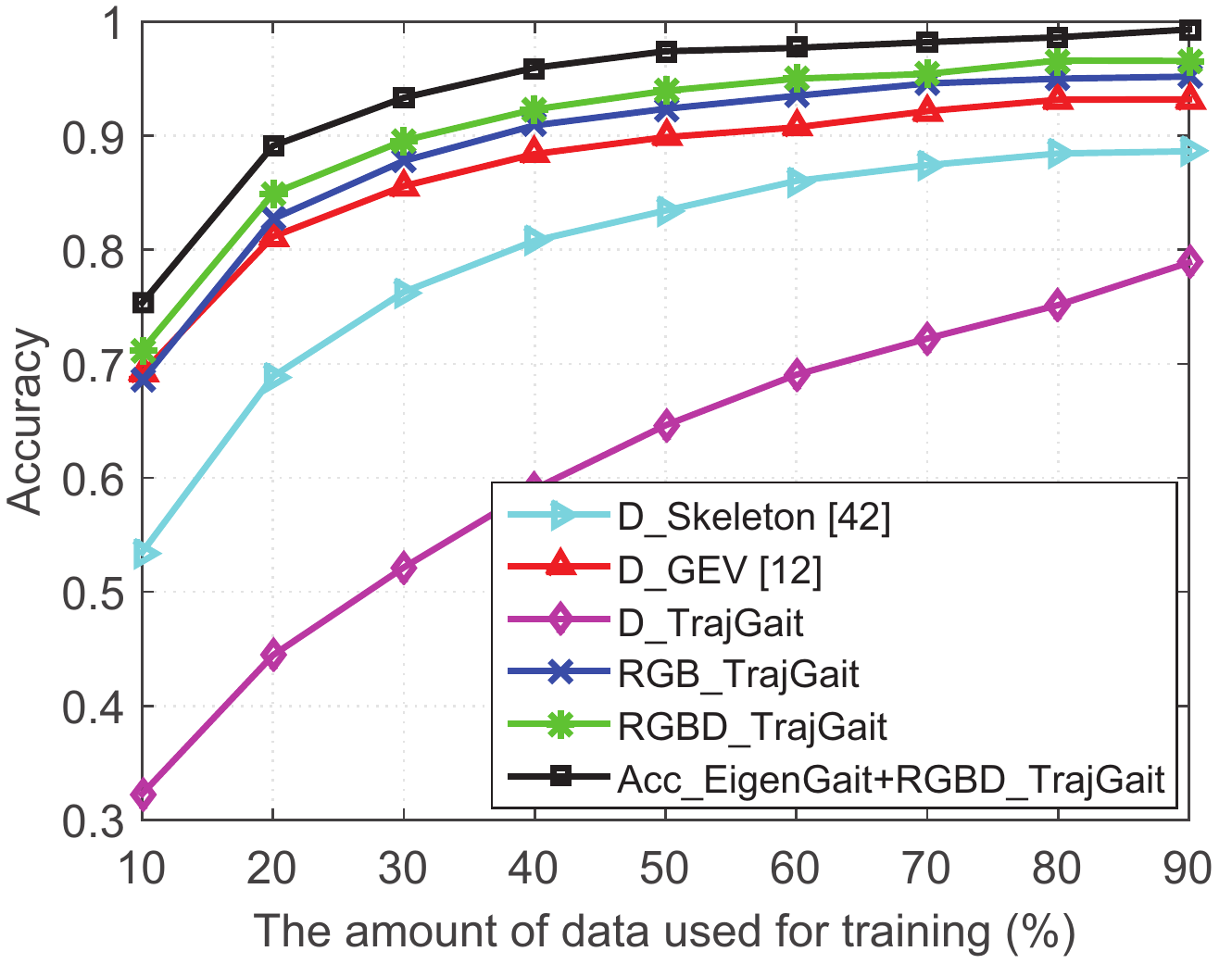}}\hspace{-1mm}
    \subfigure[]{\label{Fig.sub50.2}\includegraphics[width=0.33\linewidth]{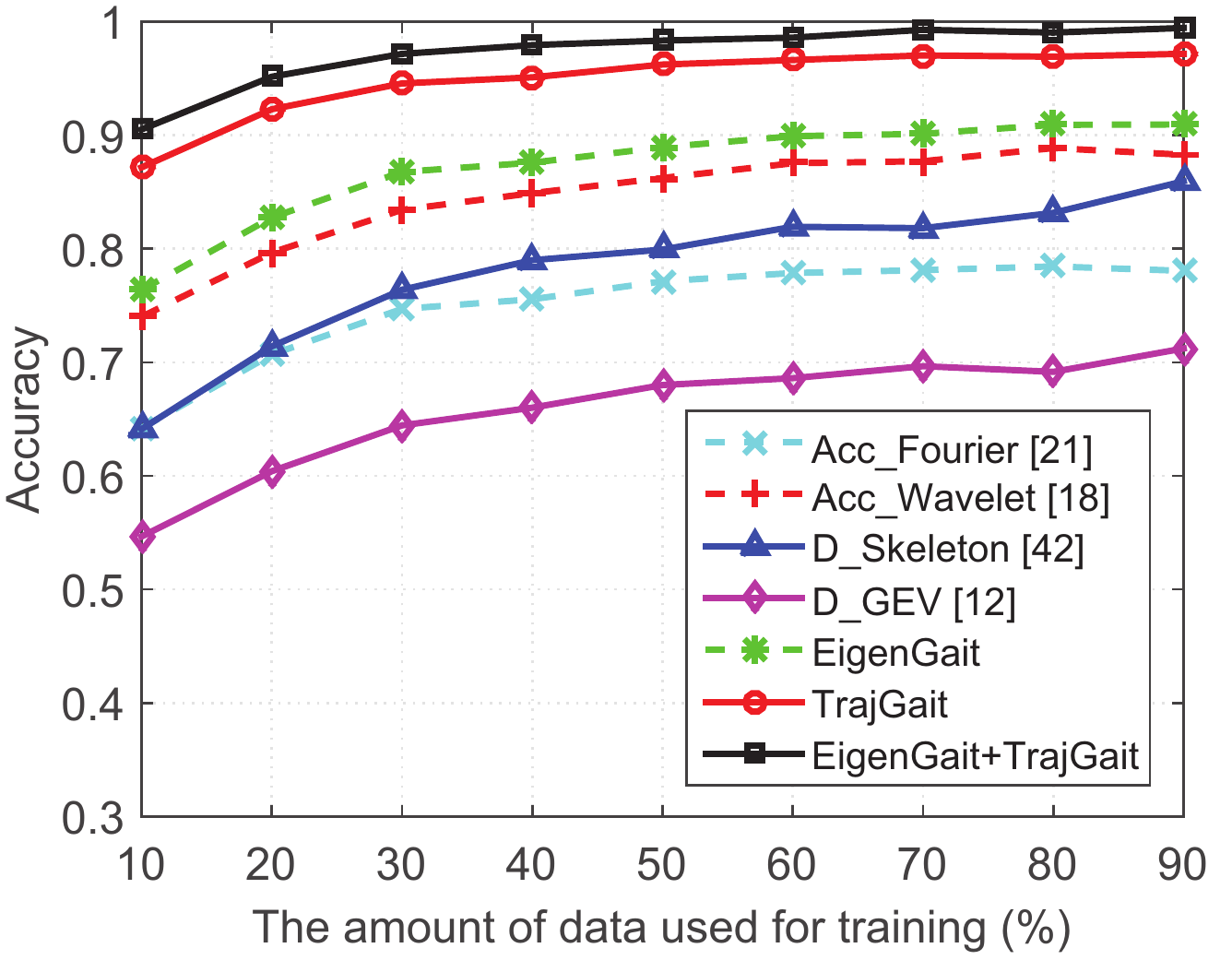}}
    \caption{{Classification performance on Dataset \#2 and Dataset \#3. (a-b) Performances on Dataset \#2, and (c) performance on the Dataset \#3.}}
    \label{fig:sub50-all}
\end{figure*}

\subsubsection{TrajGait}
We use RGBD data in Dataset \#1 to evaluate
the TrajGait algorithm. Specifically, we evaluate the TrajGait under
different $\mathcal{K}$ for K-means clustering, and linear and
non-linear SVMs. In K-means clustering, 1,000 trajectories
are randomly selected for each training sample. In feature quantization,
all trajectories of each data sample are used, which may span from about 8,000
to 15,000 in our experiments.  Figure~\ref{fig:eigen5steps}(d) shows the TrajGait accuracy
when $\mathcal{K}$=256, 512 and 1024., respectively. We can see that
the TrajGait achieves classification
accuracies higher than 0.98 when using over 20\% data for training.
A higher performance can be achieved with a larger $\mathcal{K}$,
i.e., the size of the codebook. It can also be observed that, under the
same $\mathcal{K}$, a non-linear SVM produces a little higher accuracies than
the linear one. Considering that linear SVM performs better in EigenGait and has lower
computation cost, we choose the linear SVM in the proposed gait recognition
by combining the EigenGait and TrajGait features.

\subsection{Accuracy}

We evaluate the overall performance of the proposed method, i.e.,
EigenGait+TrajGait, by comparing it with several other inertia-based
methods, color and depth based methods. Specifically, the following
methods are included in the comparison,

\begin{enumerate}[ $\vcenter{\hbox{\tiny$\bullet$}}$]
\item Acc\_Fourier~\cite{Sun14sensors}: An autocorrelation operation is first applied to the acceleration data, which is then
converted into the frequency domain using FFT. The top half of
the coefficients are selected as the gait features.

\item Acc\_Wavelet~\cite{XuCMU12btas}: The Mexican Hat Wavelet transform is
used to analyze the gait patterns from the acceleration data.

\item Acc\_EigenGait: The proposed EigenGait algorithm handles the acceleration data.

\item D\_Skeleton~\cite{Munsell12eccvw}: The position matrix on 20 joints are decomposed
by SVD, and the resulting 220-dimensional vectors are used for
gait representation.

\item D\_GEV~\cite{Sivapalan11ijcb}: The GEV is computed on the human masks extracted from depth data. The
principal component analysis is then performed the same way as in our
EigenGait for gait features.

\item D\_TrajGait: The displacement of a trajectory is calculated only
on the depth channel, with a codebook size $\mathcal{K}$=1024.

\item RGB\_TrajGait: The displacement of a trajectory is calculated
on the RGB channels, with $\mathcal{K}$=1024.

\item RGBD\_TrajGait: The full TrajGait algorithm, i.e., trajectories extracted from the RGBD
channels, with $\mathcal{K}$=1024.

\item Acc\_EigenGait+RGBD\_TrajGait: The full version of the proposed method
by combining EigenGait and TrajGait features. We normalize the EigenGait feature
and TrajGait feature independently before concatenating them together. Afterwards,
we normalize the concatenated feature as an input data for SVM.
The normalization is performed using an L1-norm measure.
\end{enumerate}

For clarity, we use Figs.~\ref{fig:sub50-all}(a) and (b) to
show the results of the acceleration-based methods and
the RGBD-based methods, respectively. In Fig.~\ref{fig:sub50-all}(a),
the proposed EigenGait is observed with a clear higher performance
than the wavelet-based or FFT-based methods, in handling
acceleration data. In Fig.~\ref{fig:sub50-all}(b) we can see that,
RGBD\_TrajGait obtains an accuracy over 0.90 when using 30\% data
for training, which is much higher than that of D\_Skeleton and
D\_GEV. The TrajGait has a higher performance on the RGB channels
than on the depth channel, which indicates that the color is more
effective than the depth in representing gait sub-dynamics.
Meanwhile, RGBD\_TrajGait outperforms
RGB\_TrajGait and D\_TrajGait, which simply demonstrates that the
color information and depth information can complement each other
in characterizing the gait. It can also be seen from
Fig.~\ref{fig:sub50-all}(b) that, a boosted performance can be achieved
by fusing EigenGait (handling acceleration data) and TrajGait
(handling RGBD data) features, i.e., EigenGait+TrajGait,
which validates the effectiveness of the proposed
multi-sensor data fusion strategy.

\subsection{Robustness}

We evaluate the robustness of the proposed method with Dataset \#3,
which contains 2,400 data samples of 50 subjects, under 8
hard-covariate conditions, as introduced in
Section~\ref{sec:dataset}. Figure~\ref{fig:sub50-all}(c) shows the
results of the proposed method and the comparison methods. We
can see that, the TrajGait+EigenGait, the
TrajGait, and the EigenGait achieves the top three performances
among all the methods. The proposed method, i.e.,
TrajGait+EigenGait, stably hold an classification accuracy over 0.90
when varying the amount of training data from 10\% to 90\%, which
indicates the proposed method can better handle these hard covariates.

\begin{figure}[t!]
    \centering
    {\hspace{2mm}\includegraphics[width=1.07\linewidth]{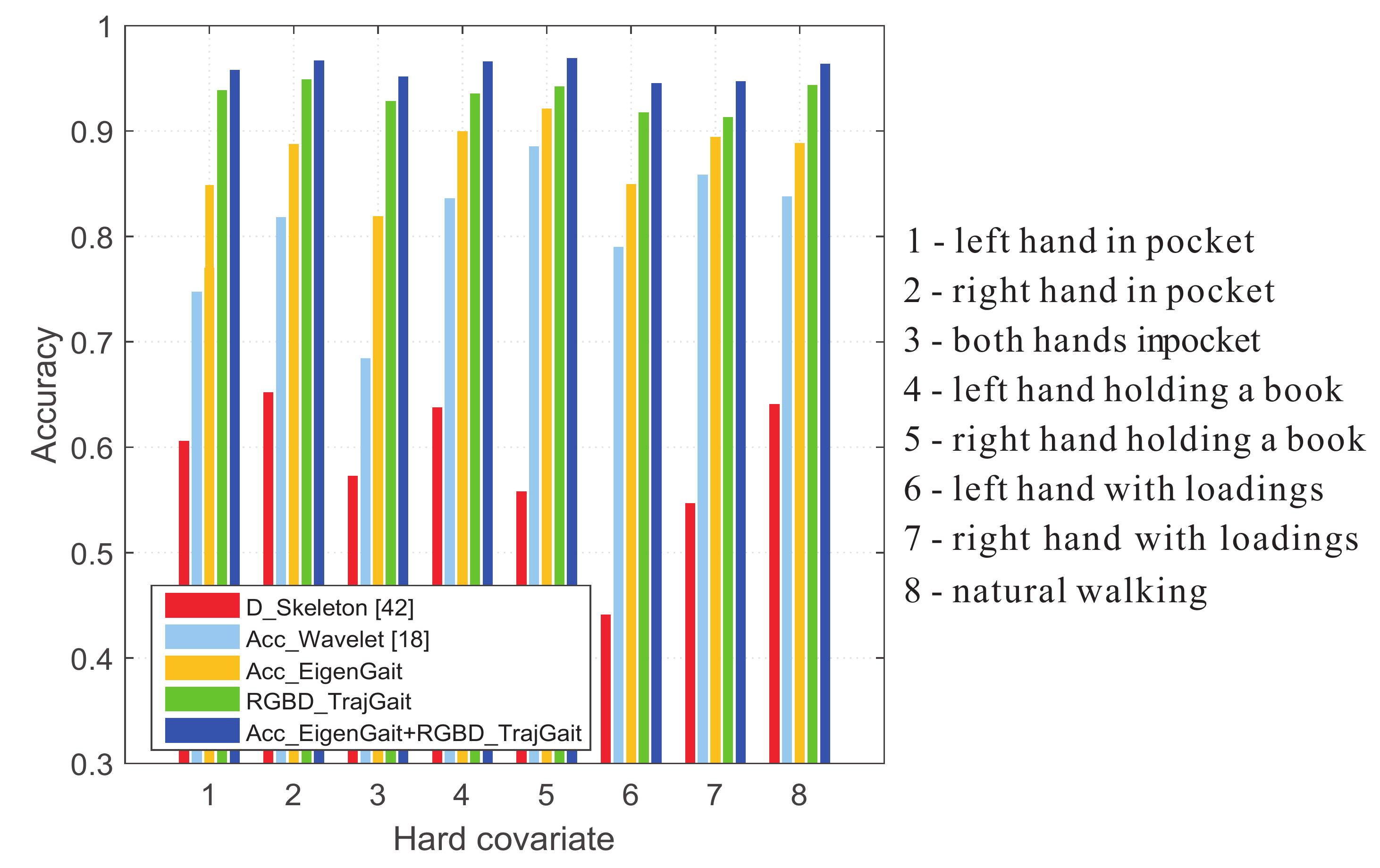}}\vspace{-0.5mm}
    \caption{{Classification accuracies under 8 hard-covariate conditions using
    20\% of the data samples for training.}}
    \label{fig:model8-2}
\end{figure}

Moreover, we investigate the detailed performance of the proposed
method by figuring out the classification accuracies on each kind of
hard covariate. As shown in Fig.~\ref{fig:model8-2}, for EigenGait,
the hard covariate `both hands in pocket' leads to the
lowest accuracy. It is because that, the acceleration would heavily
vary from normal when a person walks with both hands in the
pockets. While for TrajGait, `a hand with loadings'
will increase the difficulty for gait recognition. This is because
the loadings may bring unexpected motions in the color space, as
well as in the depth space, e.g., a bag is used to carry the
loadings in our case. For the skeleton-based and wavelet-based methods,
the average classification accuracy is about 30\% and 10\% lower than the proposed method,
respectively. Comparing with the turbulent performances of the comparison methods on different
hard covariates, the proposed method performs rather stably.

\subsection{Person-Identification Performance}

Finally, we evaluate the proposed method in the application scenario of person identification, as shown in Fig.~\ref{fig:usage}.
Half of the data samples in Dataset \#3 are used for training, and the remaining half are used for querying and identification.
The average ROC curve~\cite{fawcett2006roc,zou2014painting} is employed for performance evaluation.
For each subject, an ROC curve is computed on the results of a one-vs-all binary classification.
The ROC curve is created by plotting the true positive rate (TPR) against the false positive rate (FPR) at varying threshold settings.
The TPR and FPR are defined by

\begin{figure}[!t]
    \centering
    {\includegraphics[width=0.73\linewidth]{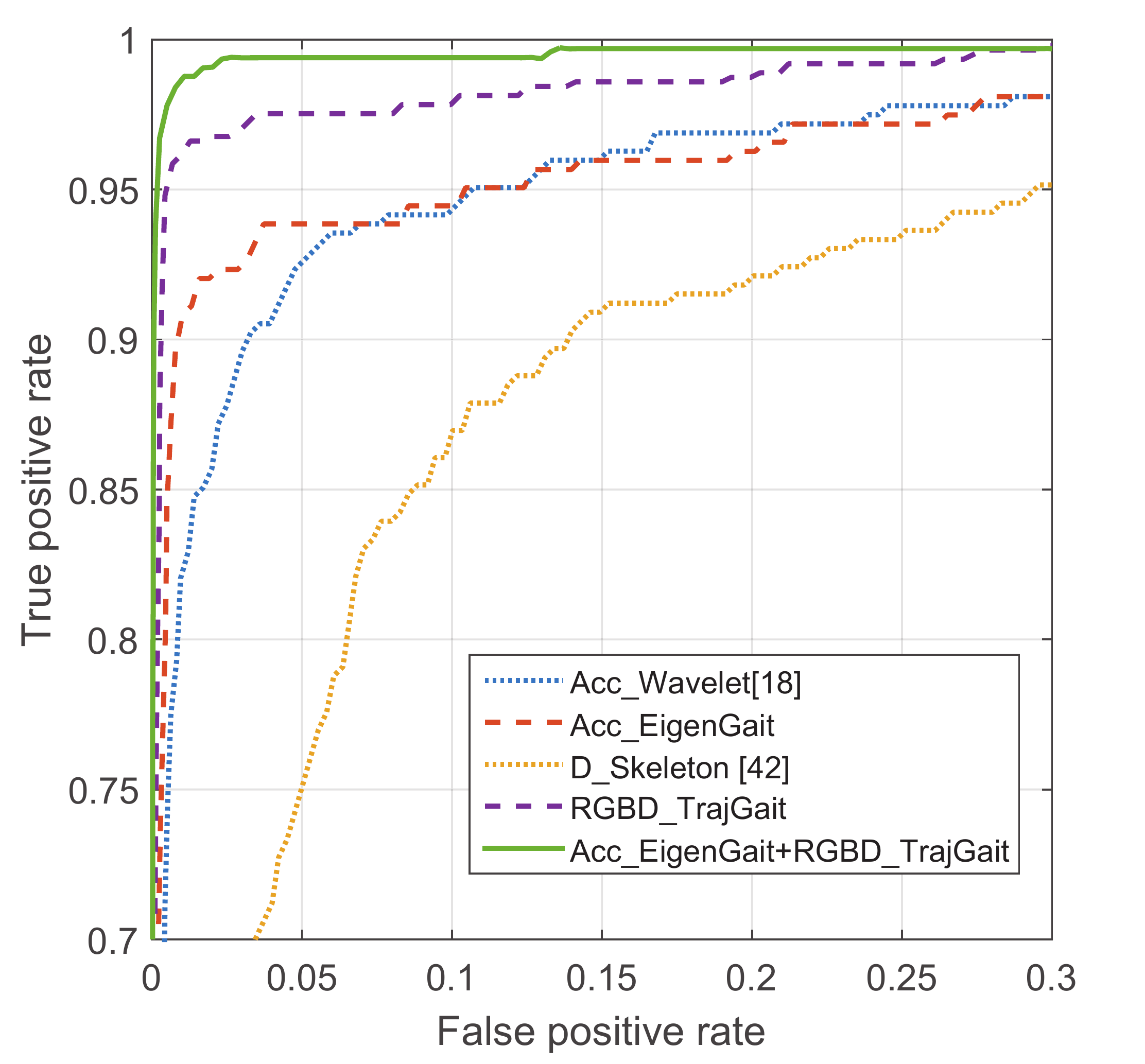}}\vspace{-0.8mm}
    \caption{{ROC curves of person identification on Dataset \#3 using 50\% data samples for training.}}
    \label{fig:model8-auth}
\end{figure}

{\small
\begin{equation}\label{eq:exp-tpr}
TPR = \frac{True\ Positive}{True\ Positive +\ False\ Negative},
\end{equation}}\vspace{-2mm}
{\small
\begin{equation}\label{eq:exp-fpr}
FPR = \frac{False\ Positive}{False\ Positive +\ True\ Negative}.
\end{equation}}
Then the average ROC curve is computed based on all ROC curves of 50 subjects.
The larger the area under the ROC curve, the better the person-identification performance.
The average ROC curves for the proposed method and the comparison methods are plotted in
Fig.~\ref{fig:model8-auth}. We can see that, the proposed method by combining
EigenGait and TrajGait achieves the best performance.
In addition, the EigenGait and the Wavelet-based method
produce competing performance, but the former achieves higher TPR than the later when the FPR is below 0.05.
Thus, the EigenGait would outperform the Wavelet-based method since a lower FPR is often required in a strict
identification system.
It can also be observed from Fig.~\ref{fig:model8-auth} that, the TrajGait uniformly outperforms the
EigenGait which may simply indicate that the TrajGait features are more discriminative by describing the detailed gait sub-dynamics.

\vspace{-2mm}
\section{Conclusion} \label{sec:conclusion}

In this paper, the inertia, color and depth sensors were integrated for accurate gait recognition and
robust person identification. Specifically, the accelerometer of smart phone and the RGBD sensor of
Kinect were employed for data collection. An EigenGait algorithm was proposed to
process the acceleration data from inertial sensor in the eigenspace,
and capture the general dynamics of the gait. A TrajGait algorithm was proposed to extract
gait features on the dense 3D trajectories from the RGBD data, and
capture the more detailed sub-dynamics. The extracted general
dynamics and detailed sub-dynamics were fused and fed into a linear
SVM for training and testing. Datasets collected from 50 subjects were used for experiments and the results
showed the effectiveness of the proposed method against several existing state-of-the-art gait recognition methods.

In the experiments, there are several other interesting findings. First, for the acceleration-based
gait recognition, the walking pace has a potential influence on accuracy
of the system. Uniform walking pace under a normal
speed produces better gait recognition than mixed walking paces. Second,
for the RGBD-based gait recognition, motion can be better captured by
wearing textured clothes, with which we can more accurately infer the
detailed gait sub-dynamics for gait recognition. Third, the proposed construction and
encoding of the 3D dense trajectories can provide more discriminative and robust gait features
under different hard-covariate conditions than other sparse joint-based trajectories.

In the future, we plan to further enhance the gait recognition
system by configuring more sensors and building more effective classifiers.
For example, more Kinects may be installed to capture multiple views of a walking person.
For the classifier, other proved techniques in classification, such as the
fuzzy-reasoning strategies~\cite{liang2000tfuzzy,li2015tac,li2015tfuzzy}
may be integrated into SVM to improve the recognition accuracy and robustness.


%
%
%

%
%

\bibliographystyle{IEEEtran}
\bibliography{refs}

\begin{thebibliography}{10}
\providecommand{\url}[1]{#1}
\csname url@samestyle\endcsname
\providecommand{\newblock}{\relax}
\providecommand{\bibinfo}[2]{#2}
\providecommand{\BIBentrySTDinterwordspacing}{\spaceskip=0pt\relax}
\providecommand{\BIBentryALTinterwordstretchfactor}{4}
\providecommand{\BIBentryALTinterwordspacing}{\spaceskip=\fontdimen2\font plus
\BIBentryALTinterwordstretchfactor\fontdimen3\font minus
  \fontdimen4\font\relax}
\providecommand{\BIBforeignlanguage}[2]{{%
\expandafter\ifx\csname l@#1\endcsname\relax
\typeout{** WARNING: IEEEtran.bst: No hyphenation pattern has been}%
\typeout{** loaded for the language `#1'. Using the pattern for}%
\typeout{** the default language instead.}%
\else
\language=\csname l@#1\endcsname
\fi
#2}}
\providecommand{\BIBdecl}{\relax}
\BIBdecl

\bibitem{Nixon2006}
M.~Nixon, T.~Tan, and R.~Chellapppa, ``Human identification based on gait,''
  \emph{Springer Science + Business Media Inc.}, 2006, ch. 1.

\bibitem{zhang2010tsmc}
J.~Zhang, J.~Pu, C.~Chen, and R.~Fleischer, ``Low-resolution gait
  recognition,'' \emph{IEEE Transactions on Systems, Man, and Cybernetics, Part
  B: Cybernetics}, vol.~40, no.~4, pp. 986--996, 2010.

\bibitem{ding2015tc}
M.~Ding and G.~Fan, ``Multilayer joint gait-pose manifolds for human gait
  motion modeling,'' \emph{IEEE Transactions on Cybernetics}, vol.~45, no.~11,
  pp. 2413--2424, 2015.

\bibitem{Kale02fg}
A.~Kale, A.~Rajagopalan, N.~Cuntoor, and V.~Krueger, ``Gait-based recognition
  of humans using continuous {HMMs},'' in \emph{International Conference on
  Automatic Face and Gesture Recognition}, 2002, pp. 336--341.

\bibitem{Han03cvpr}
J.~Han and B.~Bhanu, ``Individual recognition using gait energy image,'' in
  \emph{IEEE Conference on Computer Vision and Pattern Recognition}, 2003, pp.
  1--8.

\bibitem{Liu06pami}
Z.~Liu and S.~Sarkar, ``Improved gait recognition by gait dynamics
  normalization,'' \emph{IEEE Transactions on Pattern Analysis and Machine
  Intelligence}, vol.~28, no.~6, pp. 863--876, 2006.

\bibitem{ran2010tsmc}
Y.~Ran, Q.~Zheng, R.~Chellappa, and T.~M. Strat, ``Applications of a simple
  characterization of human gait in surveillance,'' \emph{IEEE Transactions on
  Systems, Man, and Cybernetics, Part B: Cybernetics}, vol.~40, no.~4, pp.
  1009--1020, 2010.

\bibitem{gu2010tsmc}
J.~Gu, X.~Ding, S.~Wang, and Y.~Wu, ``Action and gait recognition from
  recovered 3-d human joints,'' \emph{IEEE Transactions on Systems, Man, and
  Cybernetics, Part B: Cybernetics}, vol.~40, no.~4, pp. 1021--1033, 2010.

\bibitem{Kusa13tifs}
W.~Kusakunniran, Q.~Wu, J.~Zhang, Y.~Ma, and H.~Li, ``A new view-invariant
  feature for cross-view gait recognition,'' \emph{IEEE Transactions on
  Information Forensics and Security}, vol.~8, no.~10, pp. 1642--1653, 2013.

\bibitem{Boulgouris13tip}
N.~Boulgouris and X.~Huang, ``Gait recognition using {HMMs} and dual
  discriminative observations for sub-dynamics analysis,'' \emph{IEEE
  Transactions on Image Processing}, vol.~22, no.~9, pp. 3636--3647, 2013.

\bibitem{goffredo2010tsmc}
M.~Goffredo, I.~Bouchrika, J.~N. Carter, and M.~S. Nixon, ``Self-calibrating
  view-invariant gait biometrics,'' \emph{IEEE Transactions on Systems, Man,
  and Cybernetics, Part B: Cybernetics}, vol.~40, no.~4, pp. 997--1008, 2010.

\bibitem{Sivapalan11ijcb}
S.~Sivapalan, D.~Chen, S.~Denman, S.~Sridharan, and C.~Fookes, ``Gait energy
  volumes and frontal gait recognition using depth images,'' in
  \emph{International Joint Conference on Biometrics}, 2011.

\bibitem{John13icip}
V.~John, G.~Englebienne, and B.~Krose, ``Person re-identification using
  height-based gait in colour depth camera,'' in \emph{International Conference
  on Image Processing}, 2013, pp. 3345--3349.

\bibitem{Chatto14jvcir}
P.~Chattopadhyay, A.~Roy, S.~Sural, and J.~Mukhopadhyay, ``Pose depth volume
  extraction from rgb-d streams for frontal gait recognition,'' \emph{Journal
  of Visual Communication and Image Representation}, vol.~25, no.~1, pp.
  53--63, 2014.

\bibitem{Mantyjarvi05iccasp}
J.~Mantyjarvi, M.~Lindholm, E.~Vildjiounaite, S.~Makela, and H.~Ailisto,
  ``Identifying users of portable devices from gait pattern with
  accelerometers,'' in \emph{IEEE Conference on Acoustics, Speech, and Signal
  Processing}, 2005, pp. 973--976.

\bibitem{Liu07icbb}
R.~Liu, Z.~Duan, J.~Zhou, and M.~Liu, ``Identification of individual walking
  patterns using gait acceleration,'' in \emph{International Conference on
  Bioinformatics and Biomedical Engineering}, 2007, pp. 543--546.

\bibitem{Kwapisz10icb}
J.~Kwapisz, G.~Weiss, and S.~Moore, ``Cell phone-based biometric
  identification,'' in \emph{International Conference on Biometrics: Theory
  Applications and Systems}, 2010, pp. 1--7.

\bibitem{XuCMU12btas}
F.~Xu, C.~Bhagavatula, A.~Jaech, U.~Prasad, and M.~Savvides, ``Gait-{ID} on the
  move: Pace independent human identification using cell phone accelerometer
  dynamics,'' in \emph{International Conference on Biometrics: Theory
  Applications and Systems}, 2012, pp. 8--15.

\bibitem{Trung11ijcb}
N.~Trung, Y.~Makihara, H.~Nagahara, R.~Sagawa, Y.~Mukaigawa, and Y.~Yagi,
  ``Phase registration in a gallery improving gait authentication,'' in
  \emph{International Joint Conference on Biometrics}, 2011.

\bibitem{Trung12icb}
N.~Trung, Y.~Makihara, H.~Nagahara, Y.~Mukaigawa, and Y.~Yagi, ``Performance
  evaluation of gait recognition using the largest inertial sensor-based gait
  database,'' in \emph{IAPR International Conference on Biometrics}, 2012, pp.
  360--366.

\bibitem{Sun14sensors}
B.~Sun, Y.~Wang, and J.~Banda, ``Gait characteristic analysis and
  identification based on the {iPhone's} accelerometer and gyrometer,''
  \emph{Sensors}, vol.~14, no.~9, pp. 17\,037--17\,054, 2014.

\bibitem{Cutting77}
J.~Cutting and L.~Kozlowski, ``Recognition of friends by their walk,''
  \emph{Bulletin of the Psychonomic Society}, vol.~9, pp. 353--356, 1977.

\bibitem{Herran14sensors}
A.~M. de-la Herran, B.~Garcia-Zapirain, and A.~Mendez-Zorrilla, ``Gait analysis
  methods: an overview of wearable and non-wearable systems, highlighting
  clinical applications,'' \emph{Sensors}, vol.~14, no.~2, pp. 3362--3394,
  2014.

\bibitem{BenAbdelkader02fg}
C.~BenAbdelkader, R.~Cutler, and L.~Davis, ``Stride and cadence as a biometric
  in automatic person identification and verification,'' in \emph{International
  Conference on Automatic Face and Gesture Recognition}, 2002, pp. 372--377.

\bibitem{wang03pami}
L.~Wang, T.~Tan, H.~Ning, and W.~Hu, ``Silhouette analysis based gait
  recognition for human identification,'' \emph{IEEE Transactions on Pattern
  Analysis and Machine Intelligence}, vol.~25, no.~12, pp. 1505--1518, 2003.

\bibitem{Kale04tip}
A.~Kale, A.~Sundaresan, A.~Rajagopalan, N.~Cuntoor, A.~Roy-Chowdhury,
  V.~Kruger, and R.~Chellappa, ``Identification of humans using gait,''
  \emph{IEEE Transactions on Image Processing}, vol.~13, no.~9, pp. 1163--1173,
  2004.

\bibitem{Tao07pami}
D.~Tao, X.~Li, X.~Wu, and S.~Maybank, ``General tensor discriminant analysis
  and gabor features for gait recognition,'' \emph{IEEE Transactions on Pattern
  Analysis and Machine Intelligence}, vol.~29, no.~10, pp. 1700--1715, 2007.

\bibitem{wang03iccv}
L.~Wang, H.~Z. Ning, T.~N. Tan, and W.~M. Hu, ``Fusion of static and dynamic
  body biometrics for gait recognition,'' in \emph{International Conference on
  Computer Vision}, 2003, pp. 1449--1454.

\bibitem{Tolliver03avb}
D.~Tolliver and R.~Collins, ``Gait shape estimation for identification,'' in
  \emph{Audio- and Video-Based Biometric Person Authentication}, 2003, pp.
  734--742.

\bibitem{Han04cvpr}
J.~Han and B.~Bhanu, ``Statistical feature fusion for gait-based human
  recognition,'' in \emph{IEEE Conference on Computer Vision and Pattern
  Recognition}, 2004, pp. 842--847.

\bibitem{Zhang05neurcomp}
Z.~Zhang and N.~Troje, ``View-independent person identification from human
  gait,'' \emph{Neurocomputing}, vol.~69, no. 1-3, pp. 250--256, 2005.

\bibitem{BenAbdelkader01AVPA}
C.~BenAbdelkader, R.~Culter, H.~Nanda, and L.~Davis, ``Eigengait: Motion-based
  recognition people using image self-similarity,'' in \emph{International
  Conference on Audio and Video-based Person Authentication}, 2001, pp.
  284--294.

\bibitem{chen2011pr}
C.~Chen, J.~Liang, and X.~Zhu, ``Gait recognition based on improved dynamic
  bayesian networks,'' \emph{Pattern Recognition}, vol.~44, no.~4, pp.
  988--995, 2011.

\bibitem{venkat2011ijcv}
I.~Venkat and P.~De~Wilde, ``Robust gait recognition by learning and exploiting
  sub-gait characteristics,'' \emph{International Journal of Computer Vision},
  vol.~91, no.~1, pp. 7--23, 2011.

\bibitem{Hu13tc}
M.~Hu, Y.~Wang, Z.~Zhang, D.~Zhang, and J.~Little, ``Incremental learning for
  video-based gait recognition with {LBP} flow,'' \emph{IEEE Transactions on
  Cybernetics}, vol.~43, no.~1, pp. 77--89, 2013.

\bibitem{lam2011pr}
T.~H. Lam, K.~H. Cheung, and J.~N. Liu, ``Gait flow image: A silhouette-based
  gait representation for human identification,'' \emph{Pattern recognition},
  vol.~44, no.~4, pp. 973--987, 2011.

\bibitem{Castro14icpr}
F.~Castro, M.~Marin-Jimenez, and R.~Medina-Carnicer, ``Pyramidal fisher motion
  for multiview gait recognition,'' in \emph{International Conference on
  Pattern Recognition}, 2014, pp. 1--4.

\bibitem{Kusa14ivc}
W.~Kusakunniran, ``Attribute-based learning for gait recognition using
  spatio-temporal interest points,'' \emph{Image and Vision Computing},
  vol.~32, no.~12, pp. 1117--1126, 2014.

\bibitem{yam2002accv}
C.~Yam, M.~S. Nixon, and J.~N. Carter, ``Gait recognition by walking and
  running: a model-based approach,'' in \emph{Proceeding of the Asian
  Conference on Computer Vision}, 2002.

\bibitem{cunado2003cviu}
D.~Cunado, M.~S. Nixon, and J.~N. Carter, ``Automatic extraction and
  description of human gait models for recognition purposes,'' \emph{Computer
  Vision and Image Understanding}, vol.~90, no.~1, pp. 1--41, 2003.

\bibitem{felz2010pbmPami}
P.~F. Felzenszwalb, R.~B. Girshick, D.~McAllester, and D.~Ramanan, ``Object
  detection with discriminatively trained part-based models,'' \emph{IEEE
  Transactions on Pattern Analysis and Machine Intelligence}, vol.~32, no.~9,
  pp. 1627--1645, 2010.

\bibitem{Munsell12eccvw}
B.~Munsell, A.~Temlyakov, C.~Qu, and S.~Wang, ``Person identification using
  full-body motion and anthropometric biometrics from kinect videos,'' in
  \emph{European Conference on Computer Vision Workshop}, 2012, pp. 91--100.

\bibitem{Gabel12IEMBC}
M.~Gabel, R.~Gilad-Bachrach, E.~Renshaw, and A.~Schuster, ``Full body gait
  analysis with kinect,'' in \emph{International Conference of the IEEE
  Engineering in Medicine and Biology Society}, 2012, pp. 1964--1967.

\bibitem{Igual13eurasip}
L.~Igual, A.~Lapedriza, and R.~Borras, ``Robust gait-based gender
  classification using depth cameras,'' \emph{EURASIP Journal on Image and
  Video Processing}, pp. 1--11, 2013.

\bibitem{chattop2014tifs}
P.~Chattopadhyay, S.~Sural, and J.~Mukherjee, ``Frontal gait recognition from
  incomplete sequences using rgb-d camera,'' \emph{IEEE Transactions on
  Information Forensics and Security}, vol.~9, no.~11, pp. 1843--1856, 2014.

\bibitem{Derawi10iih}
M.~Derawi, P.~Bours, and K.~Holien, ``Improved cycle detection for
  accelerometer based gait authentication,'' in \emph{International Conference
  on Intelligent Information Hiding and Multimedia Signal Processing}, 2010,
  pp. 312--317.

\bibitem{Gafurov10icaina}
D.~Gafurov, E.~Snekkenes, and P.~Bours, ``Improved gait recognition performance
  using cycle matching,'' in \emph{International Conference on Advanced
  Information Networking and Applications}, 2010, pp. 836--841.

\bibitem{Chan11icpct}
H.~Chan, H.~Zheng, H.~Wang, R.~Gawley, M.~Yang, and R.~Sterritt, ``Feasibility
  study on iphone accelerometer for gait detection,'' in \emph{International
  Conference on Pervasive Computing Technologies for
  Healthcare(PervasiveHealth)}, 2011, pp. 184--187.

\bibitem{Ngo14pr}
T.~Ngo, Y.~Makihara, H.~Nagahara, Y.~Mukaigawa, and Y.~Yagi, ``The largest
  inertial sensor-based gait database and performance evaluation of gait-based
  personal authentication,'' \emph{Pattern Recognition}, vol.~47, no.~1, pp.
  228--237, 2014.

\bibitem{zhang2015tc}
Y.~Zhang, G.~Pan, K.~Jia, M.~Lu, Y.~Wang, and Z.~Wu, ``Accelerometer-based gait
  recognition by sparse representation of signature points with clusters,''
  \emph{IEEE Transactions on Cybernetics}, vol.~45, no.~9, pp. 1864--1875,
  2015.

\bibitem{kobayashi2004action}
T.~Kobayashi and N.~Otsu, ``Action and simultaneous multiple-person
  identification using cubic higher-order local auto-correlation,'' in
  \emph{International Conference on Pattern Recognition}, 2004, pp. 741--744.

\bibitem{gkalelis2009icip}
N.~Gkalelis, A.~Tefas, and I.~Pitas, ``Human identification from human
  movements,'' in \emph{IEEE International Conference on Image Processing},
  2009, pp. 2585--2588.

\bibitem{iosifidis2012tifs}
A.~Iosifidis, A.~Tefas, and I.~Pitas, ``Activity-based person identification
  using fuzzy representation and discriminant learning,'' \emph{IEEE
  Transactions on Information Forensics and Security}, vol.~7, no.~2, pp.
  530--542, 2012.

\bibitem{iosifidis2013action}
A.~Iosifidis, A.~Tefas, and I.~Pitas, ``Person identification from actions based on dynemes and discriminant
  learning,'' in \emph{IEEE International Workshop on Biometrics and
  Forensics}, 2013, pp. 1--4.

\bibitem{lu2012mm}
J.~Lu, J.~Hu, X.~Zhou, and Y.~Shang, ``Activity-based person identification
  using sparse coding and discriminative metric learning,'' in \emph{ACM
  international conference on Multimedia}, 2012, pp. 1061--1064.

\bibitem{yan2014eccvw}
H.~Yan, J.~Lu, and X.~Zhou, ``Activity-based person identification using
  discriminative sparse projections and orthogonal ensemble metric learning,''
  in \emph{European Conference on Computer Vision Workshops}, 2014, pp.
  809--824.

\bibitem{yan2016neuralcomp}
H.~Yan, ``Discriminative sparse projections for activity-based person
  recognition,'' \emph{Neurocomputing}, 2016.

\bibitem{wu2014cvprw}
J.~Wu, J.~Konrad, and P.~Ishwar, ``The value of multiple viewpoints in
  gesture-based user authentication,'' in \emph{IEEE Conference on Computer
  Vision and Pattern Recognition Workshops}, 2014, pp. 90--97.

\bibitem{kvi2015CVPR}
I.~Kviatkovsky, I.~Shimshoni, and E.~Rivlin, ``Person identification from
  action styles,'' in \emph{IEEE Conference on Computer Vision and Pattern
  Recognition Workshops}, 2015, pp. 84--92.

\bibitem{turk1991eigenfaces}
M.~Turk and A.~Pentland, ``Eigenfaces for recognition,'' \emph{Journal of
  cognitive neuroscience}, vol.~3, no.~1, pp. 71--86, 1991.

\bibitem{farneback2003two}
G.~Farneb{\"a}ck, ``Two-frame motion estimation based on polynomial
  expansion,'' in \emph{Image Analysis}.\hskip 1em plus 0.5em minus 0.4em\relax
  Springer, 2003, pp. 363--370.

\bibitem{Shotton13pami}
J.~Shotton, R.~Girshick, A.~Fitzgibbon, T.~Sharp, M.~Cook, M.~Finocchio,
  R.~Moore, P.~Kohli, A.~Criminisi, A.~Kipman, and A.~Blake, ``Efficient human
  pose estimation from single depth images,'' \emph{IEEE Transactions on
  Pattern Analysis and Machine Intelligence}, vol.~35, no.~12, pp. 2821--2840,
  2012.

\bibitem{wang2011cvpr}
H.~Wang, A.~Kl{\"a}ser, C.~Schmid, and C.-L. Liu, ``Action recognition by dense
  trajectories,'' in \emph{IEEE Conference on Computer Vision and Pattern
  Recognition}, 2011, pp. 3169--3176.

\bibitem{kusakunniran2012tsmc}
W.~Kusakunniran, Q.~Wu, J.~Zhang, and H.~Li, ``Gait recognition across various
  walking speeds using higher order shape configuration based on a differential
  composition model,'' \emph{IEEE Transactions on Systems, Man, and
  Cybernetics, Part B: Cybernetics}, vol.~42, pp. 1654--1668, 2012.

\bibitem{LinCJ11a}
C.~C. Chang and C.~J. Lin, ``{LIBSVM}: a library for support vector machines,''
  \emph{ACM Transactions on Intelligent Systems and Technology}, vol.~2, pp.
  27:1--27:27, 2011.

\bibitem{fawcett2006roc}
T.~Fawcett, ``An introduction to roc analysis,'' \emph{Pattern recognition
  letters}, vol.~27, no.~8, pp. 861--874, 2006.

\bibitem{zou2014painting}
Q.~Zou, Y.~Cao, Q.~Li, C.~Huang, and S.~Wang, ``Chronological classification of
  ancient paintings using appearance and shape features,'' \emph{Pattern
  Recognition Letters}, vol.~49, pp. 146--154, 2014.

\bibitem{liang2000tfuzzy}
Q.~Liang and J.~M. Mendel, ``Interval type-2 fuzzy logic systems: theory and
  design,'' \emph{IEEE Transactions on Fuzzy Systems}, vol.~8, no.~5, pp.
  535--550, 2000.

\bibitem{li2015tac}
H.~Li, Y.~Gao, P.~Shi, and H.-K. Lam, ``Observer-based fault detection for
  nonlinear systems with sensor fault and limited communication capacity,''
  \emph{IEEE Transactions on Automatic Control}, 2015.

\bibitem{li2015tfuzzy}
H.~Li, C.~Wu, S.~Yin, and H.-K. Lam, ``Observer-based fuzzy control for
  nonlinear networked systems under unmeasurable premise variables,''
  \emph{IEEE Transactions on Fuzzy Systems}, 2015.

\end{thebibliography}

\end{document}